\pdfoutput=1

\documentclass[11pt]{article}

\usepackage{amsmath}
\usepackage{ACL2023}

\usepackage{times}
\usepackage{latexsym}
\usepackage[T1]{fontenc}

\usepackage[utf8]{inputenc}

\usepackage{microtype}

\usepackage{inconsolata}

\usepackage[capitalize]{cleveref}
\usepackage{graphicx}
\usepackage{float}
\usepackage{subcaption}
\usepackage{adjustbox}
\usepackage{booktabs}
\usepackage{multirow}
\usepackage{comment}
\usepackage[english]{babel}
\usepackage{xspace}
\usepackage{latexsym}
\usepackage{xcolor}
\usepackage{enumitem}

\newcommand{\blue}[1]{{\textcolor{blue}{#1}}}
\newcommand{\green}[1]{{\textcolor{magenta}{#1}}}
\newcommand{\xlmr}{\texttt{xlm-roberta-base}\xspace}

\makeatletter
\newcommand*{\centerfloat}{%
  \parindent \z@
  \leftskip \z@ \@plus 1fil \@minus \textwidth
  \rightskip\leftskip
  \parfillskip \z@skip}
\makeatother

\makeatletter

\newcommand{\refappendix}[1]{\cref{#1}}

\makeatletter

\newcommand\Autoref[1]{\@first@ref#1,@}
\def\@throw@dot#1.#2@{#1}
\def\@set@refname#1{
    \edef\@tmp{\getrefbykeydefault{#1}{anchor}{}}%
    \xdef\@tmp{\expandafter\@throw@dot\@tmp.@}%
    \ltx@IfUndefined{\@tmp autorefnameplural}%
         {\def\@refname{\@nameuse{\@tmp autorefname}s}}%
         {\def\@refname{\@nameuse{\@tmp autorefnameplural}}}%
}
\def\@first@ref#1,#2{%
  \ifx#2@\autoref{#1}\let\@nextref\@gobble
  \else%
    \@set@refname{#1}
    \@refname~\ref{#1}
    \let\@nextref\@next@ref
  \fi%
  \@nextref#2%
}
\def\@next@ref#1,#2{%
   \ifx#2@ and~\ref{#1}\let\@nextref\@gobble
   \else, \ref{#1}
   \fi%
   \@nextref#2%
}

\makeatother

\addto\extrasenglish{  
    
}
\addto\extrasenglish{  
    
}
\addto\extrasenglish{  
    
}

\newcommand{\newrev}[1]{#1}

\title{Analysing Cross-Lingual Transfer in Low-Resourced \\ African Named Entity Recognition}

\author{Michael Beukman \and Manuel Fokam \\
        School of Computer Science and Applied Mathematics, \\ University of the Witwatersrand, Johannesburg, South Africa \\ \texttt{\{mcbeukman,arnolfokam23\}@gmail.com}}

\begin{document}

  \maketitle
  
  \begin{abstract}
      Transfer learning has led to large gains in performance for nearly all NLP tasks while making downstream models easier and faster to train. This has also been extended to low-resourced languages, with some success. We investigate the properties of cross-lingual transfer learning between ten low-resourced languages, from the perspective of a named entity recognition task. We specifically investigate how much adaptive fine-tuning and the choice of transfer language affect zero-shot transfer performance. We find that models that perform well on a single language often do so at the expense of generalising to others, while models with the best generalisation to other languages suffer in individual language performance. Furthermore, the amount of data overlap between the source and target datasets is a better predictor of transfer performance than either the geographical or genetic distance between the languages.\footnote{We publicly release our code and models at \url{https://github.com/Michael-Beukman/NerTransfer}.}
    \end{abstract}
    
  \section{Introduction}
    The technique of using a pre-trained Natural Language Processing (NLP) model and fine-tuning it on task-specific data has recently taken the NLP world by storm, achieving state-of-the-art scores in many different tasks~\citep{soa_best, t5_model, math_dataset_soa}. Although much of the focus of pre-trained models is on English~\citep{radford2018improving_gpt,bert_og}, there are also monolingual models for other languages~\citep{devries2019bertje, canete2020spanish} and multilingual models that were trained on a large multilingual corpus~\citep{xlm_roberta, mt5}.

    Generally, the training data of these models mostly consists of higher-resourced languages (i.e., those that have large amounts of available data, such as English and German). This can result in a large discrepancy between the performance of these models on higher-resourced and low-resourced languages (where data is scarce; e.g., many African languages~\citep{alabi2022Adapting}). 

    A common challenge that arises when working with these models is the lack of task-specific data for the target language~\citep{masakhaNER}. Despite this, in many cases, we have access to data from other languages. This presents an opportunity to leverage \textit{cross-lingual transfer}, training a model on the language that we have data for and using it to make predictions for the target language. This is a common scenario, especially for low-resourced languages~\citep{masakhaNER}.

    Given this opportunity for cross-lingual transfer and the prevalence of pre-trained models, research has begun investigating the properties of these models more deeply. Studies have looked into multilingualism~\citep{pires2019How,cross_lingual_transfer_bert}, syntactic transfer~\citep{syntactic_knowledge_transfer}, and the effect of linguistic features~\citep{Dolicki2021Analysing} and other attributes~\citep{choosing_transfer_languages} on transfer performance. Despite this, it is not always clear which language we should transfer from, or which factors affect transfer~\citep{choosing_transfer_languages}.
      
    Inspired by this line of work, we focus on investigating cross-lingual transfer more deeply, specifically in a low-resourced setting. We achieve this by studying the effect of different training schemes and identifying features that are indicative of high transfer performance. We build upon the work of \citet{masakhaNER}, who recently introduced a high-quality named entity recognition dataset for ten low-resourced African languages. They also performed some analysis into which pre-trained models perform best and preliminary work into the cross-lingual transfer capabilities of models. 
    
    Our results show that adaptively fine-tuning a multilingual model on unlabelled monolingual data can improve performance on the target language, while often diminishing transfer performance by overfitting to this language. This effect is exacerbated if the monolingual dataset is large.
    Furthermore, we find that when the source and target dataset contain many shared tokens, then transfer performance is generally higher. In particular, the number of overlapping tokens between datasets is a stronger predictor of transfer performance than many other features, including the geographic distance between where the languages are spoken, and the genealogical distance between the languages.

        \section{Background and Related Work}
    
        \subsection{Named Entity Recognition (NER)}
        Named Entity Recognition is a token classification task in which the objective is to classify each token (or word) as one of a few classes, person, location, date, organisation, or no entity. NER is an impactful field~\citep{sang2003introduction_conll, arches_for_ner} with many applications~\citep{marrero2013named}, including information retrieval and spell-checking~\citep{masakhaNER}. In NER, performance is predominantly measured using the F1 score~\citep{sang2003introduction_conll,masakhaNER}, which balances precision and recall. 
        \subsection{Transfer Learning}
        Transfer learning is a technique that is often used in NLP to improve performance while requiring less task-specific data~\citep{ruder2019transfer}. 
        In one common form of transfer, we start by training a large language model on a massive corpus of unlabelled data, using these learned weights as the starting point for a specific problem, and fine-tuning further on task-specific labelled data~\citep{ruder2021lmfine-tuning}. 
        This approach has become the dominant paradigm in NLP, especially for low-resourced languages, due to its high performance when fine-tuning on small datasets~\citep{masakhaNER}. 
        The idea is that the pre-training process instills knowledge into the model about how language behaves on a general level, which then does not need to be learned from scratch using the smaller amount of task-specific data~\citep{radford2018improving_gpt,devlin2018bert}.

        If the pre-training data is in a substantially different domain from the target task, we often use \textit{adaptive fine-tuning}. This fine-tunes the pre-trained model on unlabelled data in the domain of the target task using a (masked) language modelling loss~\citep{adaptive_finetuning}. 
        A related approach, \textit{language adaptive fine-tuning} (LAFT), fine-tunes a pre-trained model on unlabelled data in the target language, which can result in improved performance on the target language~\citep{madx}.
        
        \newrev{
        Recent work has also explored learning different pre-trained base models, tailored to particular languages. For instance, \citet{afriberta} pre-train a BERT-style model on less than 1GB of text from African languages, and find that this performs well on downstream tasks, compared to massively-multilingual models that were trained on much larger datasets. \citet{ogundepo2022Afriteva} extend this by pre-training a T5-based model, expanding the applications to more general sequence-to-sequence tasks such as translation. 
        Overall, these works contribute new, Africa-centric pre-trained models and provide initial benchmarks showing that these models can perform well on downstream tasks in the languages they pre-train on (after appropriate fine-tuning).
        While this is useful for advancing the field of low-resourced NLP, they generally do not deeply investigate cross-lingual transfer learning, which is the focus of our work.
        }
        
        \subsection{Analysis}
    
        While approaches such as fine-tuning and cross-lingual transfer have been empirically shown to work well, there has been a recent trend that attempts to understand these techniques more deeply. For instance, \citet{choosing_transfer_languages} focus on finding a way to choose the best language to transfer from, and develop a model that takes in a wide range of features, such as linguistic distance, entity overlap, etc., and predicts the transfer performance. \citet{Dolicki2021Analysing} also consider features relevant to transfer and find that this depends on the task -- no single feature can explain transfer performance well across tasks.
        \citet{malkin2022Balanced} instead focus on the effect of the pre-training language and find that some languages, called \textit{donors}, transfer well to others, while others, denoted \textit{recipients}, benefit from transfer.
        Other work investigates how fine-tuning a pre-trained model alters its representations of words~\citep{zero_shot_transfer}. For example, \citet{zhou2021Closer} study the effect that fine-tuning has on the representations of a multilingual model and find that this process often clusters together the representations that correspond to the same label, thereby making the classification task easier.
    
  \section{Methodology}

  The primary goal of this paper is to gain a deeper understanding of transfer learning in low-resourced settings. To achieve this, we focus on language adaptive fine-tuning and cross-lingual transfer.

  First, we investigate the effect of LAFT on transfer performance. Due to the cost of annotation, we often have more unlabelled data than labelled, task-specific data, making LAFT very applicable.

  Secondly, we examine cross-lingual transfer in order to understand which languages transfer well to others and why. This is particularly relevant in cases where data is scarce in the target language but available in other languages, a common occurrence in low-resourced NLP. Knowing which features to consider when choosing a transfer language will be immensely useful to NLP practitioners faced with the choice of transfer language~\citep{choosing_transfer_languages}.

  We therefore consider a low-resourced NER task, using the MasakhaNER dataset~\citep{masakhaNER}. We fine-tune models on this dataset, and evaluate the effect of adding LAFT and transferring from different languages in \cref{sec:language_zeroshot}.

  Next, in \cref{sec:explain:features} we investigate how much the transfer performance correlates with various language- and dataset-based features such as data overlap and linguistic distance.
  This helps us to understand which features should be considered when choosing a source language to transfer from.

  \subsection{Data}
  We consider all ten languages from the MasakhaNER dataset. We choose these languages for three reasons: firstly, they are all low-resourced compared to high-resourced languages such as English~\citep{xlm_roberta}, allowing us to study transfer learning in the important low-resourced setting. Secondly, there exists a high-quality dataset for these languages, in contrast to many other low-resourced languages. Finally, \citet{masakhaNER} already performed extensive baseline analysis on this dataset.

  Information about the languages, including family, the region where it is spoken and dataset size, is contained in \cref{tab:maintext:mner:langs}, with additional details in \refappendix{sec:appdx:dataset}. We do note that all the languages use the Latin script, except Amharic, which uses the Fidel script. Igbo, Wolof and Yorùbá use diacritics, which are symbols attached to some letters (e.g. in ``{\d e}''), which affect the pronunciation of the word.

  \begin{table*}[h]
    \centerfloat
    \caption{Language details, partially reproduced from \citet{masakhaNER}, with permission. The \textit{NER} and \textit{LAFT Size} columns contain the number of sentences in the NER training dataset and the unlabelled LAFT dataset, respectively. \textit{Country} is the top one or two countries with the most speakers of the language, from \citet{ethnologue}.}
    \label{tab:maintext:mner:langs}
    \begin{adjustbox}{width=0.9\linewidth}
      
\begin{tabular}{lllllrrr}\toprule
    \textbf{Language} & \textbf{Lang. Code}     &         \textbf{Family}               & \textbf{Country}      & \textbf{Region}         & \textbf{Speakers}           & \textbf{NER Size} & \textbf{LAFT Size}                           \\\midrule
    Amharic           & \texttt{amh}           &         Afro-Asiatic-Ethio-Semitic    & Ethiopia      & \blue{East}             & 33M                            & 1,750              & 3.1M                  \\
    Hausa             & \texttt{hau}             &         Afro-Asiatic-Chadic           & Nigeria, Niger      & \green{West}            & 63M                            & 1,903              & 3.1M                  \\
    Igbo              & \texttt{ibo}              &         Niger-Congo-Volta-Niger       & Nigeria      & \green{West}            & 27M                            & 2,233              & 1.1M                  \\
    Kinyarwanda       & \texttt{kin}       &         Niger-Congo-Bantu             & Rwanda, Uganda      & \blue{East}             & 12M                            & 2,110              & 726K                  \\
    Luganda           & \texttt{lug}           &         Niger-Congo-Bantu             & Uganda      & \blue{East}             & 7M                             & 2,003              & 506K                  \\
    Luo Nilo          & \texttt{luo}          &         Saharan                       & Kenya      & \blue{East}             & 4M                             & 644                & 160K                  \\
    Nigerian Pidgin   & \texttt{pcm}   &         English Creole                & Nigeria      & \green{West}            & 75M                            & 2,100              & 207K                  \\
    Swahili           & \texttt{swa}           &         Niger-Congo-Bantu             & Tanzania, Kenya       & Central \& \blue{East}  & 98M                            & 2,104              & 12.6M                  \\
    Wolof             & \texttt{wol}             &         Niger-Congo-Senegambia        & Senegal       & \green{West} \& NW      & 5M                             & 1,871              & 42K                  \\
    Yorùbá            & \texttt{yor}            &         Niger-Congo-Volta-Niger       & Nigeria      & \green{West}            & 42M                            & 2,124              & 910K                  \\
    \bottomrule
\end{tabular}

      \end{adjustbox}
  \end{table*}

  \section{Experiments}
  \subsection{Experimental Setup}
  Our experiments largely consist of fine-tuning pre-trained language models on NER data and evaluating their cross-lingual transfer performance.
  We perform each experiment 5 times with different random seeds and report the mean performance.
  We note that the standard deviation across the different seeds is often quite large when performing transfer, i.e., when the fine-tuning and testing language are not the same. More details are in \refappendix{sec:additional_transfer}.

  We use the MasakhaNER implementation\footnote{\url{https://github.com/masakhane-io/masakhane-ner/}} and use the same hyperparameters and language codes as \citet{masakhaNER}. All metrics reported are overall F1 scores on the test set (to compare against prior work), using the ``begin'' repair strategy as specified by \citet{palen2021seqscore}. More details regarding the training and evaluation procedures can be found in \refappendix{sec:appdx:hyperparameters}.

  \subsection{Models}
  We mainly consider two types of models, the first being \xlmr, denoted as ``base''. Secondly, we consider LAFT models, obtained by fine-tuning \xlmr on unlabelled monolingual data from a specific language. We choose to use \xlmr as the base model due to its high performance and fast training~\citep{masakhaNER}.
  This model was pre-trained on a large corpus consisting of data from 100 languages, including Amharic, Hausa and Swahili.

  We then fine-tune these models on the NER data of a specific language. For clarity, we contract the training procedure of a model, for example, $\texttt{base} \to \texttt{hau} \to \texttt{wol}$ is the \xlmr model that performed language-adaptive fine-tuning on Hausa, followed by NER fine-tuning on Wolof. More information about these models and the LAFT process is contained in \refappendix{sec:appdx:hyperparameters}.

  \section{Cross-lingual Transfer}
  \label{sec:language_zeroshot}
  Here we investigate the zero-shot transfer performance of \xlmr and the language-adaptive models. For each pair of languages $X, Y$, we take the model fine-tuned on NER data from language $X$ and evaluate its performance on language $Y$. To evaluate the effect of LAFT, we use both \texttt{base} and $\texttt{base} \to{X}$ (the latter model being obtained by performing LAFT on language $X$).

  This experiment simulates the scenario where we do not have ample labelled data in the source language, but we possess task-specific data in a different language. Here, we must choose the best language to transfer from. This is a common setup in NLP, particularly for low-resourced languages~\citep{choosing_transfer_languages,madx,masakhaNER}. Leveraging cross-lingual transfer can lead to useable models in a data-scarce setting, where no data is available for the target language.
    
  \subsection{Results} 
\begin{figure*}
    \centerfloat
    \begin{subfigure}[t]{0.45\linewidth}
      \captionsetup{width=.9\linewidth}
    \includegraphics[width=1\linewidth]{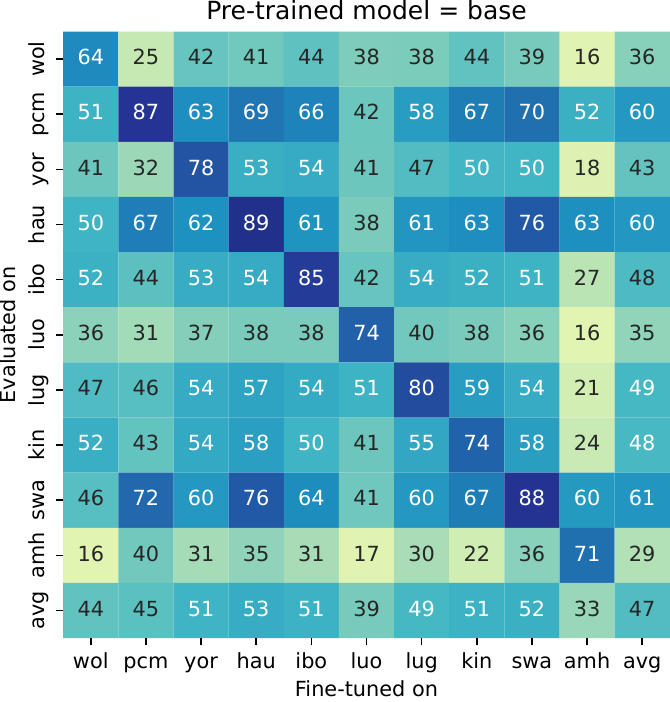}
    \caption{$\texttt{base} \to \text{X-axis}$ (no LAFT)}
    \label{fig:base_results_heatmap}
  \end{subfigure}\hfill
  \begin{subfigure}[t]{0.45\linewidth}
    \captionsetup{width=.9\linewidth}
    \includegraphics[width=1\linewidth]{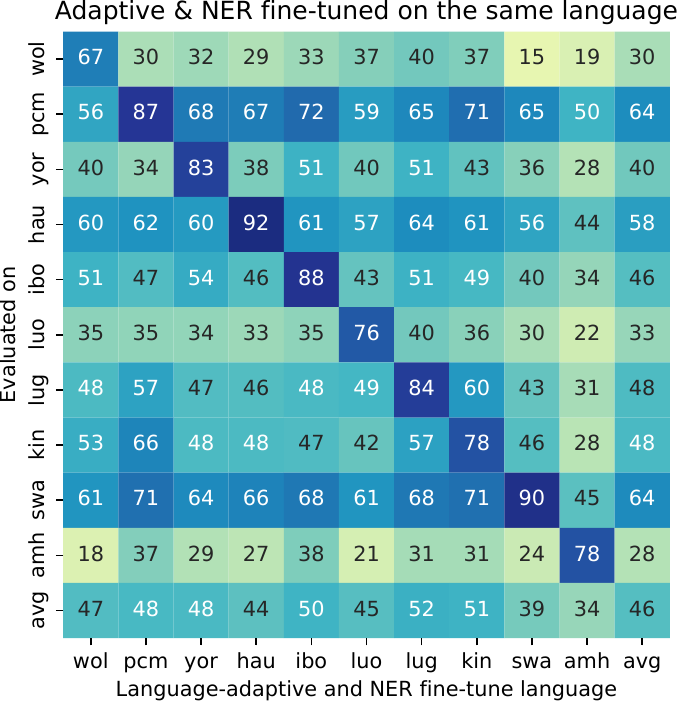}
    \caption{$\texttt{base} \to \text{X-axis} \to \text{X-axis}$}
    \label{fig:lang_results_heatmap}
  \end{subfigure}

  \caption{Heatmaps indicating the average performance over 5 seeds of specific models on specific languages (y-axis) after being fine-tuned on another language's NER data (x-axis). 
  \textit{avg} indicates the average transfer performance per row or column, respectively. This calculates the average of the entire row or column excluding the diagonal.
  }

  \label{fig:all_heatmaps}
  \end{figure*}

  These results are shown in \cref{fig:all_heatmaps}, with the y-axis representing the evaluation language, while the x-axis represents either the language we performed NER fine-tuning on (\cref{fig:base_results_heatmap}), or both the LAFT and NER fine-tuning language (\cref{fig:lang_results_heatmap}).
  \looseness=-1 In \cref{fig:base_results_heatmap}, as expected, the diagonal is brighter than the off-diagonal elements, as fine-tuning on the same language one evaluates on improves scores significantly. The best zero-shot transfer language generally performs well, obtaining 10-20 F1 lower than training on the target language. When evaluating on Yorùbá, Igbo, Luo and Amharic, however, transfer performance is significantly lower. Igbo and Yorùbá's use of diacritics, or that Amharic has a different script, may be the cause of this. Luo's low performance could be because it has a large number of entities that occur only in its dataset. In addition, $\texttt{base}$ did not train on any Luo data, and Luo is from a different family to all of the other languages.

  Furthermore, while Amharic transfers poorly on average, it transfers reasonably well to Swahili, Hausa and Nigerian Pidgin. The reason may be that Amharic, Swahili and Hausa were included in the base model's pre-training data, while Nigerian Pidgin shares many similarities with English, another pre-training language. Thus, the pre-trained model may have some link between its representations for Amharic and the other languages it jointly pre-trained on. Fine-tuning on Amharic changes these shared representations, leading to improved transfer results (see \refappendix{sec:embeddings} for details).

  \subsubsection*{Observation: LAFT on the target language improves downstream performance}
  Comparing the diagonals in \cref{fig:base_results_heatmap} and \cref{fig:lang_results_heatmap}, we can see that the LAFT models usually perform much better than the base model after subsequent NER fine-tuning.
  Of particular interest is the large improvement we see in Yorùbá and Amharic, where the language adaptive models outperform the base models by $+5$ and $+7$ F1, respectively. This could be because Yorùbá contains diacritics, and Amharic does not use the Latin script, making the language adaptive fine-tuning phase crucial to adapt the model to the specific characteristics of these languages. On average, by using language adaptive fine-tuning on the target language, we can improve the F1 performance by approximately $3$ F1 after subsequent NER fine-tuning.

  \subsubsection*{Observation: Performing LAFT on a large dataset can diminish transfer performance}
  \looseness=-1 While performing LAFT improves performance on the language we fine-tune on, transfer performance often shows a mixed result.
  For some language pairs, using a model that has been subject to language-adaptive fine-tuning on the same language as one fine-tunes on helps (e.g. the \textit{pcm} column and \textit{kin} row), but for others, this effect is minor, or even negative (e.g. \textit{yor} transferring to \textit{lug}). For some languages, notably Swahili and Hausa, using adaptively fine-tuned models (and then fine-tuning on NER data from the same language) significantly diminishes the transfer capabilities from these languages, possibly indicating overfitting. This is similar to what \citet{madx} found when performing adaptive fine-tuning on the source language -- transfer performance generally decreased. We investigate this further (more details in \refappendix{sec:overfitdataset}) and find that those languages with fewer sentences in the language adaptive datasets transfer better on average after performing LAFT and NER fine-tuning. There is a statistically significant correlation, with Pearson's $R=-0.82$, between the number of sentences in the LAFT dataset and the average improvement in transfer performance when using a LAFT model compared to using the base model. This suggests that larger LAFT datasets result in more overfitting and less transfer.

  \subsubsection*{Recommendation} 
  \begin{itemize}[noitemsep,nolistsep]
    \item Use LAFT on the target language prior to fine-tuning on NER data in the same language.
    \item If transfer performance is the priority, however, LAFT on a large dataset in the NER fine-tuning language should be avoided.
  \end{itemize}

  \section{Explaining Transfer Performance}
  \label{sec:explain:features}
  To explain some of the results shown in the previous section, here we examine other dataset and language features, determining whether they have any correlation with the transfer performance.
  \subsection{Data Overlap}
  The first feature we consider is the word overlap between the different languages' datasets. We do so because in NER, a token classification task, a model would benefit greatly from previously encountering an entity. Thus, if languages $X$ and $Y$ share tokens, a model trained on $X$ would perform well on the already-seen tokens in language $Y$.

  We call a token overlapping when the same token is labelled as the same entity type in two different datasets (e.g. \texttt{John[NAME]} would overlap with \texttt{John[NAME]}, but would not overlap with \texttt{John[ORG] Deere[ORG]}). 
  To calculate the overlap between source language $X$ and target language $Y$, we find all of the named entities (i.e., all tokens that are not labelled as ``Other'') that occur in both datasets and count the total number of times each token occurred in either dataset (e.g. if \texttt{John} occurred twice in $X$ and three times in $Y$, we count it five times). We do not distinguish between tokens that are at the beginning of an entity or in the middle thereof (i.e., we consider B-PER and I-PER to be the same for this experiment). We also consider the entire dataset, i.e., train + dev + test, to obtain a more representative sample.

  There are alternative ways to calculate overlap, such as only taking into account unique entities (which we avoid as one entity overlapping multiple times is relevant), or determining the fraction of overlapping tokens instead of the absolute number~\citep{choosing_transfer_languages}. However, we find that these alternative methods generally produce similar results and lead to similar conclusions, so the specific calculation method does not have a significant impact. In \refappendix{sec:other_overlap_methods}, we provide a more in-depth explanation of these methods and their results.

  \subsubsection{Results}
  \cref{fig:data_overlap} shows the overlap between each pair of languages, with the diagonal being proportional to the number of entities for each language.
  Wolof and Luo have much less data than the other languages, and thus much less overlap, potentially explaining why these two performed poorly in previous experiments. In particular, Wolof has around three times more sentences than Luo, but fewer entities, indicating that entities are sparsely distributed throughout its sentences. 
  Moreover, Amharic, due to it being written in a different script than all of the other languages, does not have any lexical overlap.
  Finally, there seems to be a large amount of data overlap in general, e.g., Swahili and Hausa have around $33\%$ of their tokens overlapping.

  \subsubsection*{Observation: Data Overlap Strongly Correlates with Transfer Performance}

\begin{figure*} 
      \begin{subfigure}[t]{0.45\linewidth}
        \centering\captionsetup{width=.95\linewidth}
        \resizebox{1\linewidth}{1\linewidth}{\includegraphics[width=1\linewidth]{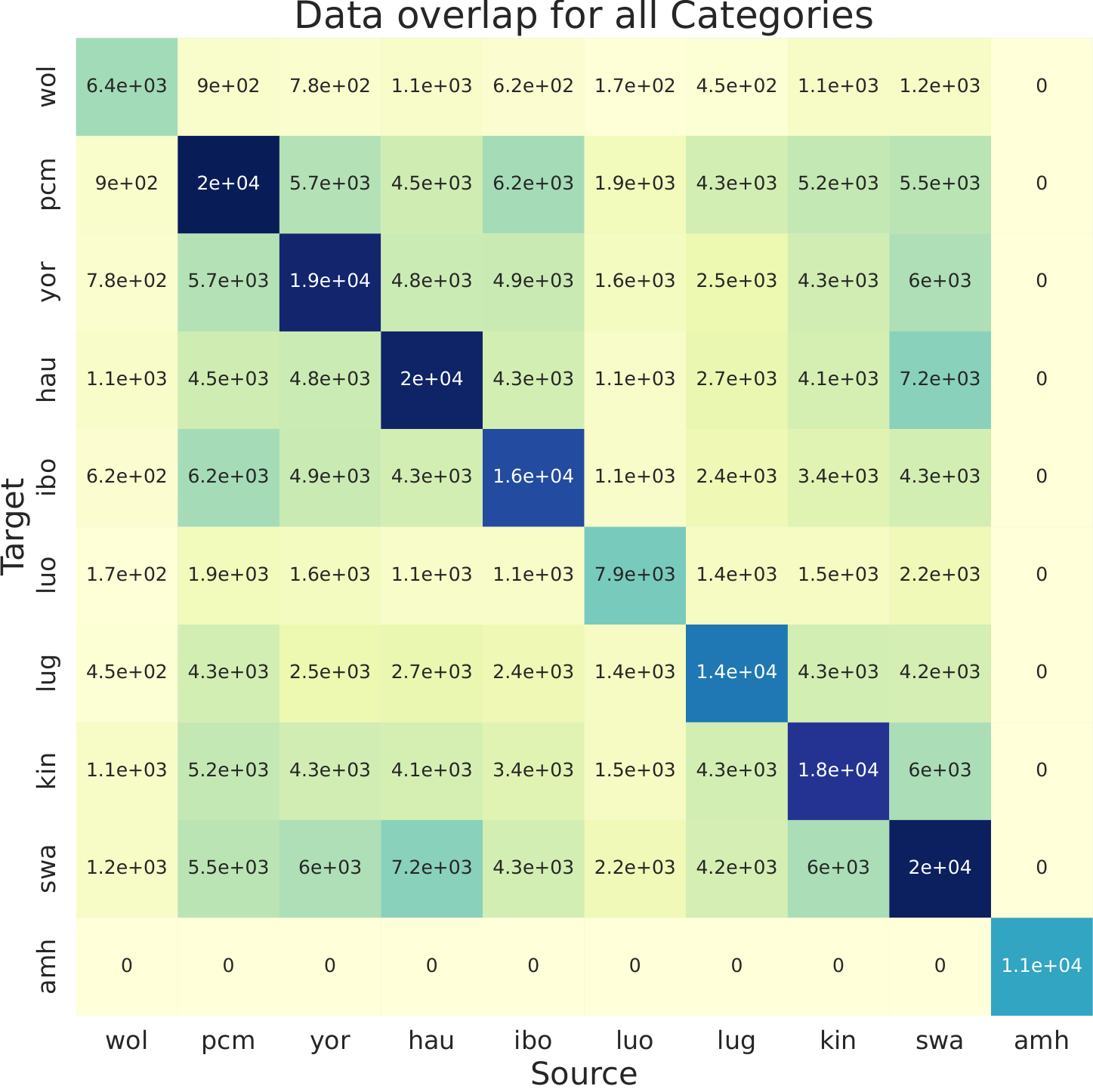}}
        \subcaption{Data Overlap. Row $i$, column $j$ indicates the overlap if $j$ was the source (i.e., fine-tuning language) and $i$ was the target (i.e., evaluation language).}
        \label{fig:data_overlap}
      \end{subfigure}\hfill
      \begin{subfigure}[t]{0.45\linewidth}
        \centering\captionsetup{width=.95\linewidth}
        \resizebox{1\linewidth}{1\linewidth}{\includegraphics[width=1\linewidth]{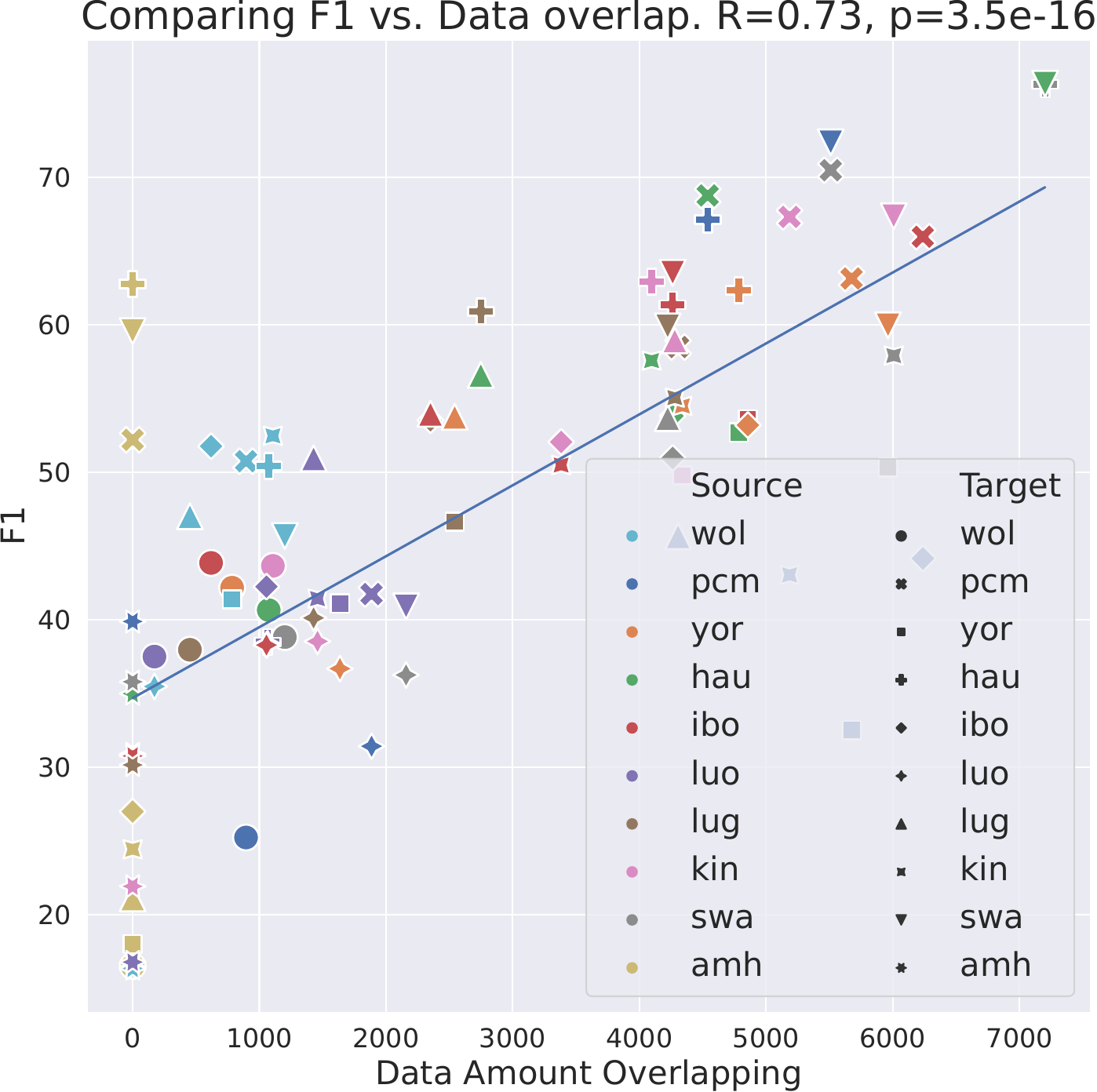}}
        \subcaption{Correlation between data overlap and F1 performance when performing zero-shot transfer. 0.73 Pearson's correlation coefficient with $p < 0.05$.}
        \label{fig:data_corr}
      \end{subfigure}
      \caption{(a) Data overlap and (b) its correlation with F1. $R$ in (b) is similar without Amharic, see \cref{sec:appdx:correlation_no_amharic}.}
      \label{fig:all_data_overlap}
    \end{figure*}

  We see a strong correlation (Pearson's $R = 0.73$) between how many tokens overlap and the performance in \cref{fig:data_corr}. The procedure here was simply to compute the correlation between the data overlap (as in \cref{fig:data_overlap}) and the performance when fine-tuning on one language and evaluating on another, starting from the pre-trained base model (as in \cref{fig:base_results_heatmap}). We do not consider the diagonal elements, as they contain the performance of evaluating on language $X$ after fine-tuning on language $X$ and are thus not considered transfer learning.
    
  These results do not imply a causal relationship, however, as previous work has shown that lexical overlap has a negligible impact on transfer performance, and word order, model depth and other attributes contribute more~\citep{pires2019How, tran2019ZeroShot,cross_lingual_transfer_bert}. This might be specific to the task under consideration, however, as other work still has shown that, for some tasks, the word and subword overlap between languages is a useful proxy for expected performance when performing cross-lingual transfer~\citep{choosing_transfer_languages}. Additionally, NER (and other token classification tasks) may be particularly sensitive to word overlap, as the classification happens on a per-word or a per-token basis. Finally, Amharic, due to its different script, has no overlap with any other language, while still displaying some transfer, indicating that more intricate mechanisms are at play.

    \subsubsection*{Observation: Most of the overlap is in English} 

    Having shown that data overlap has such a large correlation with transfer performance, we now investigate this deeper, to see which types of entities overlap. 
    To aid in this, we classify a token as ``international'' if it falls into one of the following categories: (1) place names, such as ``Africa'', ``Washington'', ``Nairobi''; (2) numbers, such as those found in dates, e.g., 2016; (3) the names of people in English, e.g. Paul, Jean; (4) punctuation marks found in the middle of entities; and (5) common words/companies such as ``December'', ``Christmas'', ``Monday'' and ``Google''. All of these are written in English. See \refappendix{sec:international_transfer_tokens} for more details about these categories.
    Over all of the entity tokens across all languages, around 35\% of these correspond to international tokens. However, when only considering the overlapping tokens, international words are the majority, around 69\%. This seems to indicate that instead of overlapping tokens representing shared words between languages, they often represent international entities written in English. This holds even when considering only the distribution of unique tokens (instead of taking into account the number of times each token occurs). In this case, international tokens make up 28\% of all tokens compared to 64\% of overlapping tokens.
    This could be a factor present mostly in news-based data such as MasakhaNER, however, as news articles often cover globally relevant topics, leading to these types of entities being shared across datasets. Our findings in this section may partially explain why \citet{masakhaNER} obtained poor performance when transferring from Wikipedia.
    Finally, we note that the correlation results (\cref{fig:data_corr}) are similar when only considering the local or international tokens, see \cref{sec:split_overlap_local_international}.

    \subsection{Additional Features}

    \label{sec:otherfeatures}
    While we primarily focus on data overlap in this work, other features may also influence transfer performance between languages. We specifically consider the features used by \citet{choosing_transfer_languages}. This includes various language-based features, such as genetic and syntactic distance, dataset-based features, such as source dataset size, as well as the geographical distance between the countries where the languages are spoken. 
    
    \subsubsection{Language-Based Features}
    The language-based features are largely based on the URIEL database of language properties~\citep{littell2017Uriel}. These features are: \emph{Genetic}, \emph{Inventory}, \emph{Syntactic} and \emph{Phonological} distance. \emph{Genetic Distance} is how different two languages are based on their language families. The other distances measure the cosine distance between vectors representing each language's syntax or phonology, derived from various linguistic databases~\citep{lewis2009ethnologue,wals}.
    \subsubsection{Dataset-Based Features}
    
    The dataset-based features are \emph{Source Dataset Size}, representing the number of sentences in the source dataset and; \emph{Source over Target Size Ratio}, the number of sentences in the source dataset divided by the number of sentences in the target dataset. We add two similar features, the number of named entities in the source dataset, and the ratio between this and the number of entities in the target dataset.
    \subsubsection{Geographical Distance}
    The geographic distance is calculated as a normalised distance between the geographic center of where the language's speakers reside~\citep{littell2017Uriel, hammarstrom2018glottolog}. This, however, is potentially problematic, especially if a language is spoken in multiple countries and is somewhat spread out. For instance, Swahili is spoken across Kenya, Tanzania, and other countries, but the ``geographic center'' for Swahili is marked as a point in Southeastern Tanzania~\citep{hammarstrom2018glottolog}. 
    As this may not be particularly accurate, we also experiment with a different approach of calculating the geographic distance between two languages, that of the shortest distance between all of the countries where each language is spoken (obtained from \citet{hammarstrom2018glottolog}). If these countries share a border, the distance is zero. However, this new method results in distances that are closely correlated with those obtained previously and result in similar conclusions, so we do not consider it further.
    Finally, \textit{Featural Distance} is the cosine distance between vectors consisting of the four language-based features and the geographical distance.
    
    \subsubsection*{Observation: Language-based features do not correlate strongly with transfer performance}
    Similarly to data overlap, we compute the correlation between these features and the transfer performance, with results in \cref{tab:langrank_features} (see \refappendix{sec:other_features} for plots). We find that most of these features exhibit a poor correlation with the transfer performance, and not all of the correlations are statistically significant. The genetic distance is the language-based feature with the highest correlation with transfer performance, at Pearson's $R=-0.30$ (i.e., closer languages tend to transfer better). This is still much weaker than the data overlap's correlation of $R=0.73$.
    Overall, this suggests that the most important feature for transfer performance is the overlap between the source and target datasets, instead of how close the languages are.
    However, it may still be best to consider several different factors instead of any single one~\citep{choosing_transfer_languages}.

    \begin{table}
      \centerfloat
      \caption{Pearson's correlation coefficient and the corresponding p-value for features used by \citet{choosing_transfer_languages}. The data overlap row is the same as in \cref{fig:all_data_overlap}. The first five features are not statistically significant, as $p \geq 0.05$.}
      \label{tab:langrank_features}
      \begin{adjustbox}{width=0.9\linewidth}
        \begin{tabular}{llrr}
\toprule
                          Feature &       Type &     R &      p \\
\midrule
                Featural Distance & Linguistic & -0.00 &      1 \\
            Phonological Distance & Linguistic & -0.02 &   0.84 \\
               Inventory Distance & Linguistic & -0.06 &   0.55 \\
    Source Over Target Size Ratio &    Dataset & -0.20 &  0.056 \\
              Geographic Distance & Geographic & -0.21 &   0.05 \\
              Source Dataset Size &    Dataset &  0.23 &  0.029 \\
               Syntactic Distance & Linguistic & -0.23 &  0.028 \\
        Source Number Of Entities &    Dataset &  0.29 & 0.0063 \\
Source Over Target Entities Ratio &    Dataset & -0.30 & 0.0044 \\
                 Genetic Distance & Linguistic & -0.30 & 0.0041 \\
\textbf{Data Overlap} & Dataset & \textbf{0.73} & $\mathbf{3.5\times 10^{-16}}$ \\
\bottomrule
\end{tabular}

        \end{adjustbox}
    \end{table}
    
    \subsubsection*{Recommendation} 
    \begin{itemize}[noitemsep,nolistsep]
      \item Choosing a source language for NER based on its data overlap with the target is promising.
      \item Other features have small correlations with transfer performance, much less than data overlap, and should not be used as a primary reason for choosing a specific language.
    \end{itemize}

  \section{Discussion}
  Our work follows a recent trend of analysing empirical results more deeply, attempting to better understand the underlying phenomena~\citep{choosing_transfer_languages,zhou2021Closer}. We specifically consider cross-lingual transfer for low-resourced languages, investigating the effect of LAFT, the choice of transfer language, and which features are indicative of high transfer performance.

  In line with recent work, we find that LAFT can improve performance on downstream tasks~\citep{masakhaNER,alabi2022Adapting}. We also discover that performing this process on a large dataset can inhibit transfer performance. This motivates having more general models instead of overspecialised ones, which would ideally be more robust. 
  \newrev{The work of \citet{alabi2022Adapting} may be particularly relevant here, as they perform LAFT on multiple source languages, and encounter less of a loss in generalisation performance compared to our case of performing LAFT on only one language.}

  We further find that data overlap between the source and target languages correlates strongly with transfer performance, and that this may provide a way to choose a promising language to transfer from. Many other language-specific features have a much lower correlation with transfer performance. This suggests that, for token classification tasks such as NER, data overlap is potentially more important than language similarities.

  This may not always be the case, however. Performance in other tasks, such as machine translation, may be less influenced by the amount of data overlap. Furthermore, the MasakhaNER dataset largely consists of annotated news articles. This type of data may skew more towards discussing international entities than, say, local history or fact-based text such as Wikipedia. In these cases, geographical or linguistic distance may contribute more to transfer than data overlap. Thus, while we highlight some important results, they may not necessarily apply to other tasks and domains. This should be investigated in future work.

  One promising avenue of investigation for future work is to examine transfer performance when all international tokens are removed, to determine if this would diminish the correlation with data overlap, resulting in other features becoming more important for transfer. Using a more sophisticated strategy than only counting overlapping words when they exactly match would be promising, potentially resulting in the identification of similar, but slightly different, words between related languages.
  Finally, while we considered ten low-resourced African languages, it would be valuable to extend this study to other languages \newrev{and regions} to determine how well our conclusions generalise.

  \section{Conclusion}
  In this paper, we conduct a thorough examination of transfer learning in low-resourced African languages, focusing on language-adaptive fine-tuning and cross-lingual transfer. We find that language-adaptive fine-tuning on a large dataset can lead to improved performance on the target language, but at the cost of reduced transfer performance.
    
  We further demonstrate that data overlap between the source and target datasets is a powerful predictor of transfer performance in NER, surpassing other factors such as geographical or genealogical distance. This, however, does not necessarily imply that data overlap is the cause of transfer performance, as Amharic, without any overlap, still displays some transfer. We also find that English words make up the bulk of the overlapping tokens.

  Ultimately, while more work is needed, we hope that our analysis could inform some of the experimental decisions and transfer considerations when dealing with lower-resourced languages, thereby improving the quality of NER models for these languages.
  \section*{Acknowledgements}
  Computations were performed using High Performance Computing infrastructure provided by the Mathematical Sciences Support unit at the University of the Witwatersrand.
  We thank the reviewers for their helpful and insightful comments, which helped to strengthen the final version of this paper.
  Finally, thanks to Devon Jarvis, Jade Abbot, Steven James and Benjamin Rosman for useful input.

  \newrev{
  \section*{Limitations}
  While we believe that our work is valuable, it has several shortcomings that could be addressed in future work. First, our focus is solely on one task---NER. While this enables us to perform detailed experiments and analysis, the disadvantage is that our results may not be general to all NLP tasks. Furthermore, as mentioned in the discussion section, our overlap results may be quite particular to NER.
  Therefore, it would be particularly promising to extend our work to other tasks, such as machine translation, sentiment classification, text classification, etc. 

  Second, our work only focuses on a subset of ten African languages. While Africa exhibits a large amount of linguistic diversity, and has several low-resourced languages, our conclusions may not necessarily be applicable to all low-resourced languages, or languages in other regions, such as Asia, Latin America, etc. It would be beneficial to extend our work to other languages and regions by using some of the more recent datasets for low-resourced languages~\citep{prabhakar2021Clneril,adelani2022Thousand,adelani2022Masakhaner,ebrahimi2021Americasnli,patil2022L3Cubemahaner}.
  Relatedly, we only considered one dataset, MasakhaNER. While this dataset is of high quality, it is also relatively small. It would be valuable to investigate whether our results hold on lower-quality datasets, as low-resourced languages often lack high-quality datasets such as the one we considered in this work.

  Finally, to isolate the training procedure, we focused only on one pre-trained model, \xlmr. Again, this enabled us to perform in-depth analysis, but it would be valuable to extend this work to other models, such as mBERT~\citep{bert_og}, AfriBERTa~\citep{afriberta} and AfriTeVa~\citep{ogundepo2022Afriteva}, to determine if our results generalise to other models.

  }
  \bibliographystyle{acl_natbib}
  
  \bibliography{bib}
  
  \newpage \appendix
  \crefalias{section}{appendix}
  
  \section*{Appendix}
  \renewcommand{\blue}[1]{{\textcolor{blue}{#1}}}
\renewcommand{\green}[1]{{\textcolor{magenta}{#1}}}
\label{sec:appdx}

\section{Dataset}\label{sec:appdx:dataset}
As mentioned in the main text, we used the MasakhaNER dataset~\citep{masakhaNER}.\footnote{Available at \url{https://github.com/masakhane-io/masakhane-ner}} Information about the dataset, including the number of sentences and data sources, broken down by language, is shown in \cref{tab:mner:data}. \citet{masakhaNER} discuss the characteristics of the languages in more depth. 

Most of the data was sourced from various news websites around the same time, with e.g. the Swahili and Hausa data both coming from the VOA website. While the authors of \citet{masakhaNER} do not know for certain whether the Hausa and Swahili data are translations of each other, it is quite likely that the events covered are similar, as the data is from around the same period.
\begin{table*}[h!]
  \centerfloat
  \caption{Information about the different data sources and breakdowns of the NER data per language. Reproduced from \citet{masakhaNER}, with permission.}
  \label{tab:mner:data}
  \begin{adjustbox}{width=1\linewidth}
  \begin{tabular}{lllcrrrrrr}\toprule
    Language Data & Source  & Train/ dev/ test & \#Anno    & PER & ORG & LOC & DATE & \% of Entities in Tokens & \#Tokens \\\midrule
    Amharic                      & DW \& BBC                  & 1750/ 250/ 500  & 4 & 730   &         403   &     1,420 &     580     & 15.13 &  37,032 \\
    Hausa                        & VOA Hausa                 &  1903/ 272/ 545  & 3 & 1,490 &         766   &     2,779 &     922     & 12.17 &  80,152  \\
    Igbo                         & BBC Igbo                  &  2233/ 319/ 638  & 6 & 1,603 &         1,292 &     1,677 &     690     & 13.15 &  61,668  \\
    Kinyarwanda                  & IGIHE news                &  2110/ 301/ 604  & 2 & 1,366 &         1,038 &     2,096 &     792     & 12.85 &  68,819  \\
    Luganda                      & BUKEDDE news              &  2003/ 200/ 401  & 3 & 1,868 &         838   &     943   &     574     & 14.81 &  46,615  \\
    Luo                          & Ramogi FM news            &  644/ 92/ 185    & 2 & 557   &         286   &     666   &     343     & 14.95 &  26,303  \\
    Nigerian Pidgin              & BBC Pidgin                &  2100/ 300/ 600  & 5 & 2,602 &         1,042 &     1,317 &     1,242   & 13.25 &  76,063  \\
    Swahili                      & VOA Swahili               &  2104/ 300/ 602  & 6 & 1,702 &         960   &     2,842 &     940     & 12.48 &  79,272  \\
    Wolof                        & Lu Defu Waxu \& Saabal     & 1871/ 267/ 536 & 2 & 731   &         245   &     836   &     206     & 6.02  &  52,872 \\
    Yorùbá                       & GV \& VON news             & 2124/ 303/ 608  & 5 & 1,039 &         835   &     1,627 &     853     & 11.57 &  83,285 \\\bottomrule
    \end{tabular}
    \end{adjustbox}
\end{table*}
\section{Hyperparameters and Reproducibility}\label{sec:appdx:hyperparameters}
We make our code available to reproduce our experiments. \cref{tab:hyperparams} contains the hyperparameters that we used when training the models. We used the same hyperparameters and base code as \citet{masakhaNER}. For all experiments, in total, we used around 1000 GPU hours, on an internal cluster. The model we use, \xlmr, has 270M parameters.
\subsection{Language Adaptive Fine-tuning Procedure}
The language adaptive models, introduced by \citet{masakhaNER}, had the following procedure: Take the \xlmr model as a starting point and fine-tune this on unlabelled, monolingual data for one language (e.g. Swahili, Wolof, etc.) using a \newrev{masked} language modelling loss. This was done separately for each language, resulting in ten separate language-adaptive models.
The data, including its source and the number of sentences, used for this process is described in Table 10 of \citet{masakhaNER}.

\subsection{Language Split of XLM-Roberta}
XLM-Roberta~\citep{xlm_roberta} is a multilingual large language model trained on a large corpus consisting of 100 languages. In particular, it was trained on 8 African languages, including 3 languages contained in the MasakhaNER dataset, Amharic, Swahili and Hausa. The amount of data for these African languages is quite small, however, consisting of 68M tokens for Amharic, 275M for Swahili and 56M for Hausa, compared to e.g. 55B for English and 10B for French and German. Thus, while the model was trained on some African languages, they only make up a small fraction of the entire pre-training dataset.

Additionally, it is important to question whether XLM-Roberta was either pre-trained or adaptively fine-tuned on, for example, the test dataset for some of the languages. \citet{masakhaNER} do not know this for certain. It is unlikely, however, as the data of XLM-Roberta was extracted from the CommonCrawl 2018 snapshot, whereas \citet{masakhaNER} created and annotated the MasakhaNER dataset in 2020 and 2021 from current (at the time) news data.

\begin{table}
  \centerfloat
  \caption{Hyperparameters for the fine-tuning experiments}
  \label{tab:hyperparams}
  \begin{tabular}{lr}\toprule
    \textbf{Hyperparameter} & \textbf{Value}\\\midrule
    Number of Seeds & 5\\
    Fine-tuning Epochs & 50\\
    Maximum Sequence Length & 200\\
    Batch Size & 32\\
    Learning Rate & 5e-5\\
    \bottomrule
    \end{tabular}
\end{table}

\subsection{NER and its evaluation}\label{sec:appdx:ner}
In Named Entity Recognition, we often have a distinction between the start of a multi-word entity, and the continuation of one. For instance, John Deere would be labelled as B-ORG I-ORG (denoting the \textbf{b}eginning and \textbf{i}nside of an entity). However, in some cases, the gold standard, ``correct'' labels often have invalid transitions, such as I-ORG being immediately after O, bypassing the required B-ORG label~\citep{palen2021seqscore}. Relatedly, the output of a model may also contain some of these invalid transitions. This complicates evaluation, which has resulted in methods being developed to correct these problems. One common approach is the ``begin'' repair method, where any invalid ``I-'' is replaced by a ``B-''~\citep{palen2021seqscore}. After the label sequence has been repaired, the standard evaluation procedure is then used, comparing the predicted output to the ground-truth annotations. We used this begin repair strategy in our experiments.

\section{Additional Results and Analysis}\label{sec:appdx:additional_results}
This section contains additional experiments and results. In particular, we consider the correlation between overfitting (i.e. transferring worse to other languages) after performing LAFT and the size of the LAFT corpus in \cref{sec:overfitdataset}. \cref{sec:more:laft:exps} contains more LAFT experiments, considering the effect of performing LAFT on a language other than the fine-tuning one. In \cref{sec:additional_transfer} we have additional transfer results, specifically considering the different NER categories in isolation, and expanding upon the increased variance we found when performing transfer. We next detail the process when classifying tokens as international in \cref{sec:international_transfer_tokens}. \cref{sec:other_overlap_methods} covers the various other overlap calculations we consider, confirming that they all lead to similar conclusions. \cref{sec:split_overlap_local_international} considers correlation results when splitting the data and performance into international and local subsets. \cref{sec:appdx:correlation_no_amharic} performs the correlation analysis between performance and data overlap without considering Amharic, which has no overlap with any other language.
\cref{sec:other_features} contains more in-depth definitions of the other features, besides data overlap, that we considered, as well as plots showing the correlation of each feature with transfer performance.
In \cref{sec:combiningdatasets} we consider the effect of training on a combination of datasets. Finally, in \cref{sec:embeddings} we consider the representations of the pre-trained models, how they are changed by fine-tuning and how this may explain some of our transfer results.
\subsection{Overfitting vs Dataset size}\label{sec:overfitdataset}
\begin{figure}[h]
  \centering\captionsetup{width=.9\linewidth}
  \includegraphics[width=1\linewidth]{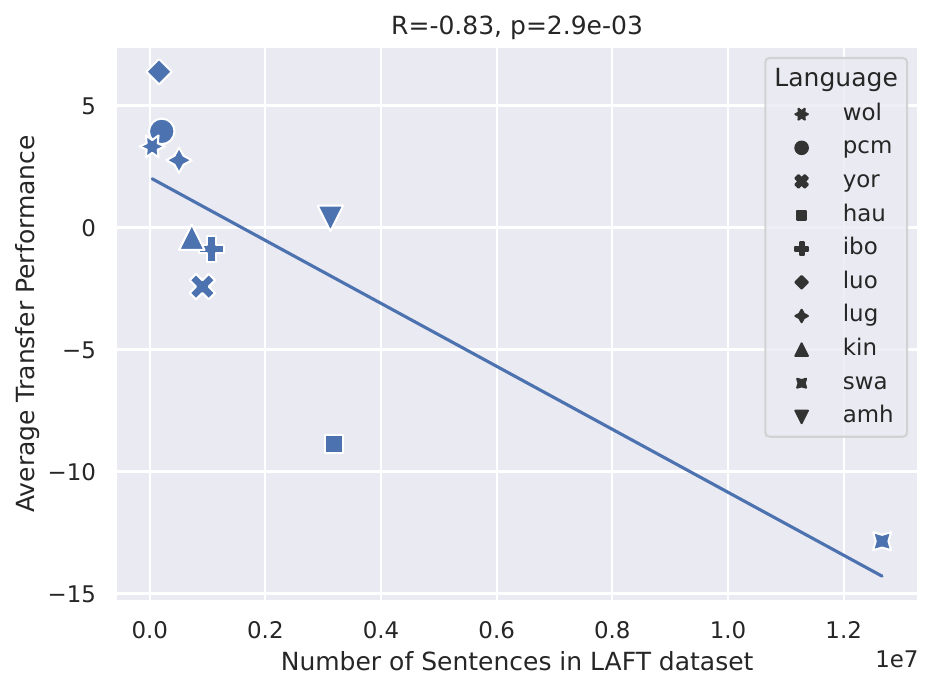}
  \caption{The correlation between the number of sentences in the LAFT sentences on the x-axis and the transfer performance delta of using this model compared to the base for fine-tuning.}
  \label{fig:overfit}
\end{figure}

In this experiment, we evaluate the effect of the size of the language adaptive dataset on the transfer performance of a model trained on downstream data. To do this, we take the (a) $\text{base} \to X \to X$ models, for each language $X$; i.e., those that performed language adaptive fine-tuning on language $X$ and additional NER fine-tuning on the same language. We then evaluate these models on the 9 other languages. We do the same for the (b) $\text{base} \to X$ models (i.e. the models that took the base pre-trained model and performed NER fine-tuning on language $X$). We subtract the average transfer performance of the (b) models from the (a) ones, to obtain the performance gain (or loss) after performing language adaptive fine-tuning. We then plot this quantity against the number of sentences in the language adaptive fine-tuning datasets (obtained from Table 10 in \citet{masakhaNER}) in \cref{fig:overfit}. We see a strong negative correlation (Pearson's $R=-0.82$) that is statistically significant ($p < 0.05$). This seems to indicate that the larger our language adaptive fine-tuning dataset is, the worse a downstream model will transfer.

Furthermore, we find that this result still holds if we omit the three languages included in \xlmr's pre-training dataset (Hausa, Swahili, Amharic), with $R=-0.89, p < 0.05$.

\newrev{\subsubsection{Pre-training Size}
In \cref{fig:lang_results_heatmap}, we find that, on average, only Swahili and Nigerian Pidgin perform better as target languages after performing LAFT on the source language. Hausa, for instance, does not. One reason for this may be the size of the \xlmr pre-training dataset: Swahili had 275M tokens, English 55B (Nigerian Pidgin is an English creole), Amharic 68M and Hausa only 56M. Pre-training on a large dataset may make the model less susceptible to generalising worse after performing LAFT on another language.}

\subsection{Additional LAFT Experiments}\label{sec:more:laft:exps}

\begin{table*}[h]
  \centerfloat
  \caption{Performance of different models after fine-tuning and evaluating on NER data. We use a Mann-Whitney U test~\citep{mann_whitney_test} as some data failed a Shapiro Wilks normality test~\citep{shapiro1965analysis}. $^*$ indicates a statistically significant difference ($p<0.05$) between the base model and the one under consideration, \textbf{bold} implies $^*$ and being the maximum per language. The leftmost column shows the model we started with before fine-tuning on language-specific NER data, while the other columns indicate the NER fine-tuning and evaluation language. For example, $\text{base} \to X$ is the language adaptive model for each column.}
  \label{tab:pretrained}
  \begin{adjustbox}{width=1\linewidth}
  \begin{tabular}{llllllllllll}
\toprule
{} &                      wol &         pcm &                      yor &                      hau &                      ibo &         luo &                      lug &                      kin &                      swa &                      amh &                      avg \\
Starting point for NER fine-tune &                          &             &                          &                          &                          &             &                          &                          &                          &                          &                          \\
\midrule
$\text{base}$                    &               64.2 (1.3) &  87.3 (0.9) &               77.9 (0.3) &               89.5 (0.4) &               84.9 (0.7) &  74.5 (1.3) &               80.2 (0.7) &               73.7 (0.7) &               87.8 (0.5) &               70.7 (1.1) &               79.1 (0.2) \\
$\text{base} \to X$              &               66.9 (1.7) &  87.1 (0.8) &  $\textbf{83.3 (0.3)}^*$ &  $\textbf{91.6 (0.4)}^*$ &  $\textbf{87.9 (0.5)}^*$ &  76.2 (1.2) &  $\textbf{84.5 (0.5)}^*$ &  $\textbf{78.3 (1.0)}^*$ &  $\textbf{89.6 (0.6)}^*$ &  $\textbf{78.2 (0.8)}^*$ &  $\textbf{82.4 (0.2)}^*$ \\
$\text{base} \to \text{swa}$     &  $\textbf{67.3 (1.3)}^*$ &  88.0 (0.8) &               78.3 (1.0) &    $\text{88.8 (0.2)}^*$ &               84.3 (0.8) &  77.2 (1.4) &    $\text{82.0 (0.5)}^*$ &               75.2 (1.0) &  $\textbf{89.6 (0.6)}^*$ &               68.9 (0.9) &    $\text{80.0 (0.5)}^*$ \\
$\text{base} \to \text{hau}$     &               66.1 (2.0) &  88.3 (1.0) &               78.7 (0.7) &  $\textbf{91.6 (0.4)}^*$ &               85.5 (0.4) &  75.2 (1.0) &               79.8 (0.6) &    $\text{72.1 (0.8)}^*$ &               87.6 (0.6) &    $\text{68.4 (0.5)}^*$ &               79.3 (0.3) \\
\bottomrule
\end{tabular}

\end{adjustbox}
\end{table*}

\subsubsection{Experiment} For each language $X$, we compare four different models: $\text{base}$, $\text{base} \to \text{Swahili}$, $\text{base} \to \text{Hausa}$ and $\text{base} \to \text{X}$, where the latter three were subject to further language-adaptive fine-tuning on their respective languages. The base model acts as a baseline that does not perform adaptive fine-tuning at all, just fine-tuning on NER data. The $\text{base} \to \text{X}$ model shows the benefits of using adaptive fine-tuning on the target language. The $\text{base} \to \text{Swahili}$ and $\text{base} \to \text{Hausa}$ models provide information on how downstream performance is affected by language-adaptive fine-tuning on a different language. We chose Swahili as it was the language with the most speakers and the largest dataset out of the 10 available ones~\citep{masakhaNER}, making it a promising language to transfer from. It is also spoken in Eastern Africa, like many of the other languages we consider. Hausa is chosen as a baseline, as another language with many speakers and a relatively large dataset. Hausa is predominantly spoken in Western Africa, in contrast to Swahili. We fine-tune each of these models on NER data and report the results when evaluated on the test set.

\subsubsection{Results}
We find that performing LAFT on Swahili outperforms the base model. When using Hausa as the LAFT language, however, we do not see any significant increase in performance compared to the base model when averaging over all languages. This may be due to the fact that the Hausa LAFT dataset is around 4 times smaller than the Swahili one.
Besides Swahili and Hausa themselves, four languages have a significant (more than the standard deviation) difference in performance between the Swahili and Hausa models, Igbo, Luo, Luganda and Kinyarwanda. Hausa only outperforms Swahili on Igbo, which is predominantly spoken in West Africa, whereas Swahili outperforms Hausa on the other three languages, largely spoken in East Africa. This suggests that the region of the source and target language has an impact on the effectiveness of LAFT.

\subsection{Additional Transfer Results}\label{sec:additional_transfer}
In \cref{fig:appdx:allheatmaps}, we show more transfer results. The first row contains transfer performance when fine-tuning on the x-axis and evaluating on the y-axis. \cref{fig:base_results_heatmap_appendix,fig:lang_results_heatmap_appendix} were contained in the main text, and \cref{fig:swa_results_heatmap} contains the results when using a $base \to \text{swa}$ language adaptive model.
The second row shows the standard deviations of the F1 score over 5 seeds for each model in the first row.
In particular, looking at \cref{fig:base_results_heatmap_std}, the results indicate that the standard deviation is generally higher when performing transfer (off-diagonal elements) compared to performing standard evaluation on the fine-tuned language (diagonal elements).  This suggests that transfer exhibits a lack of robustness to random initialisations. This effect is more pronounced when fine-tuning on Luo, as it has significantly less data than the other languages. When transferring from other languages to Amharic, the spread is higher than average, likely due to Amharic's different script.

\subsubsection{Performance varies wildly across the different entity types}
\begin{figure*}
  \includegraphics[width=1\linewidth]{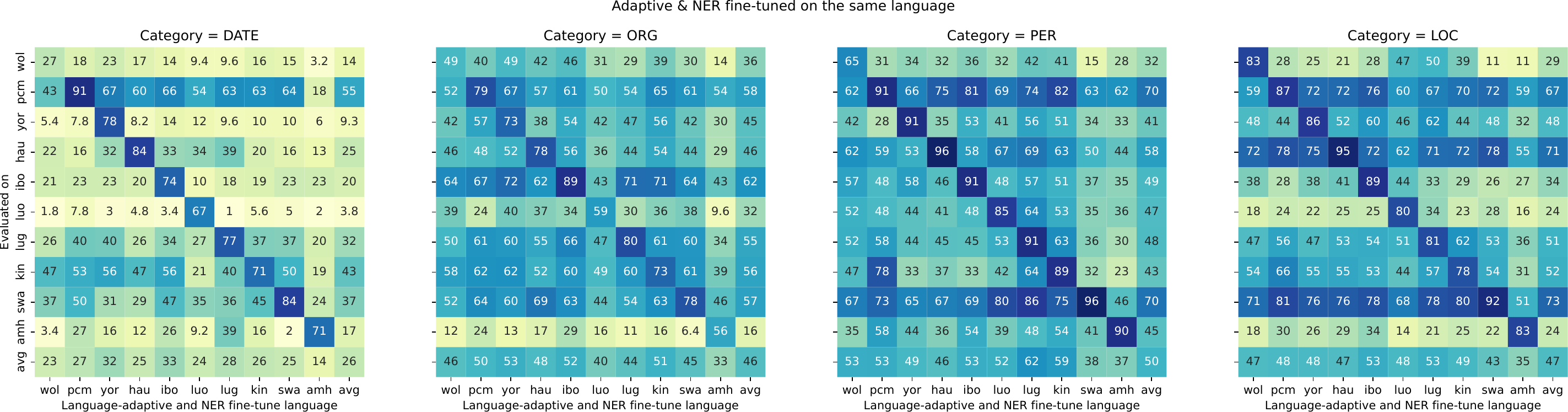}
  \caption{Heatmaps for the language-adaptive fine-tuned model (\cref{fig:lang_results_heatmap_appendix}), broken down by category.}
  \label{fig:all_heatmaps_joint}
\end{figure*}

\begin{figure*}
  \centerfloat
  \begin{subfigure}[t]{0.33\linewidth}
    \captionsetup{width=.9\linewidth}
  \includegraphics[width=1\linewidth]{all_figs/analysis_v20_results_base.pdf}
  \caption{$\text{base} \to \text{X-axis}$ (no LAFT)}
  \label{fig:base_results_heatmap_appendix}
\end{subfigure}
\begin{subfigure}[t]{0.33\linewidth}
  \captionsetup{width=.9\linewidth}
  \includegraphics[width=1\linewidth]{all_figs/analysis_v20_results_lang-specific.pdf}
  \caption{$\text{base} \to \text{X-axis} \to \text{X-axis}$}
  \label{fig:lang_results_heatmap_appendix}
\end{subfigure}
\begin{subfigure}[t]{0.33\linewidth}
  \captionsetup{width=.9\linewidth}
  \includegraphics[width=1\linewidth]{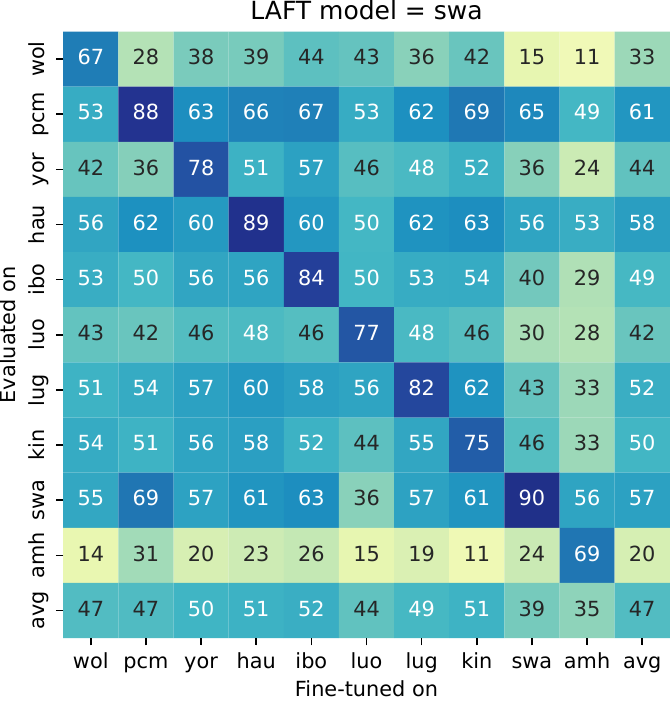}
  \caption{$\text{base} \to \text{swa} \to \text{X-axis}$}
  \label{fig:swa_results_heatmap}
\end{subfigure}

\begin{subfigure}{0.33\linewidth}
  \captionsetup{width=.9\linewidth}
  \includegraphics[width=1\linewidth]{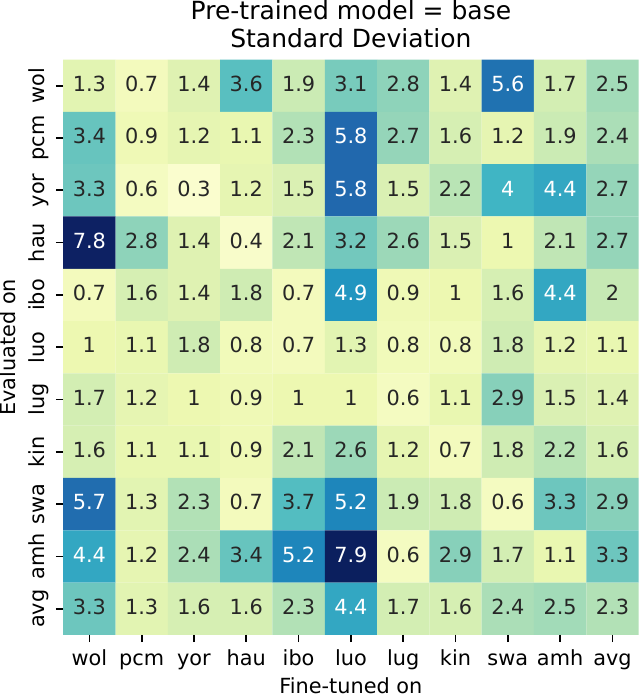}
  \caption{Standard Deviation of (a)}
  \label{fig:base_results_heatmap_std}
\end{subfigure}
\begin{subfigure}{0.33\linewidth}
  \captionsetup{width=.9\linewidth}
  \includegraphics[width=1\linewidth]{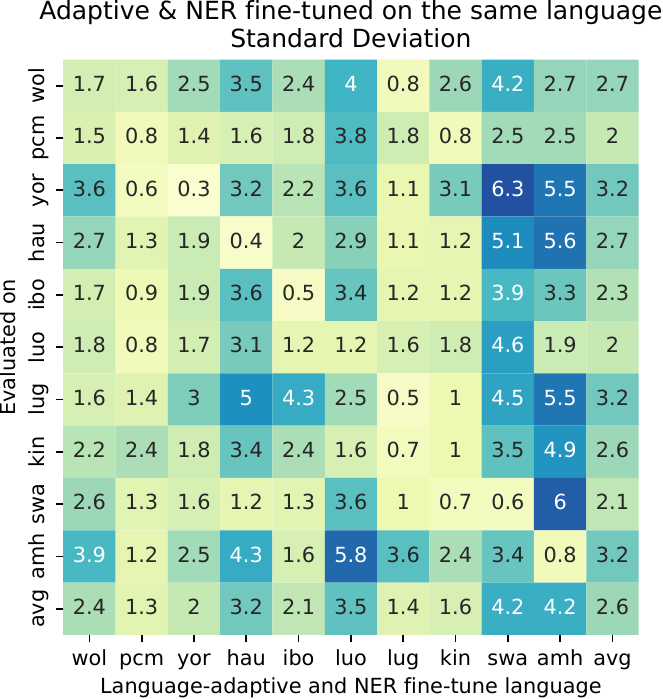}
  \caption{Standard Deviation of (b)}
  \label{fig:lang_results_heatmap_std}
\end{subfigure}
\begin{subfigure}{0.33\linewidth}
  \captionsetup{width=.9\linewidth}
  \includegraphics[width=1\linewidth]{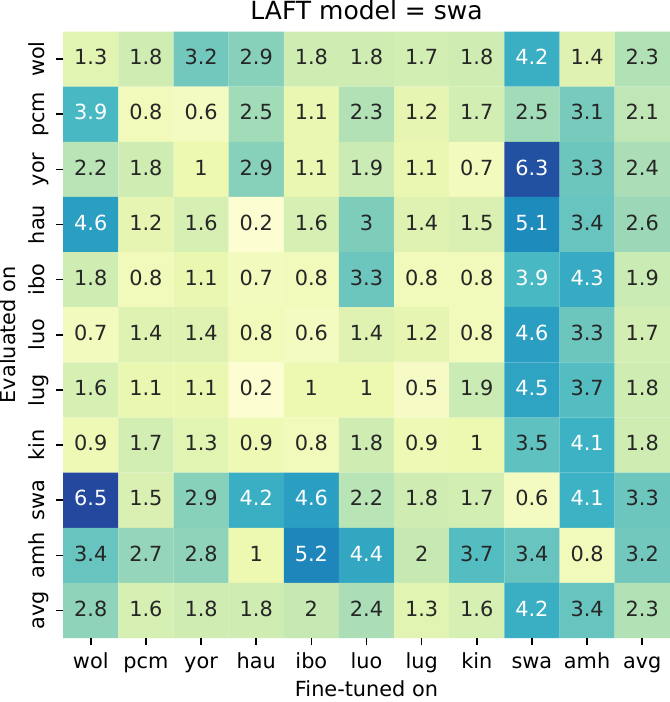}
  \caption{Standard Deviation of (c)}
  \label{fig:swa_results_heatmap_std}
\end{subfigure}

\begin{subfigure}[t]{0.33\linewidth}
    \captionsetup{width=.9\linewidth}
  \includegraphics[width=1\linewidth]{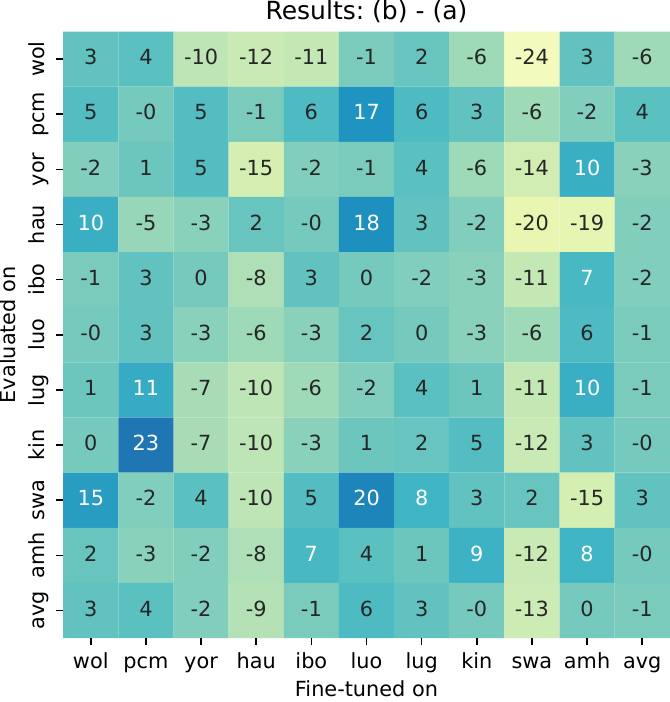}
  \caption{Performance difference when adding language-adaptive fine-tuning. Swahili and Hausa transfer worse on average, while Luo improves.}
  \label{fig:lang_results_heatmap_diff}
\end{subfigure}
\begin{subfigure}[t]{0.33\linewidth}
    \captionsetup{width=.9\linewidth}
  \includegraphics[width=1\linewidth]{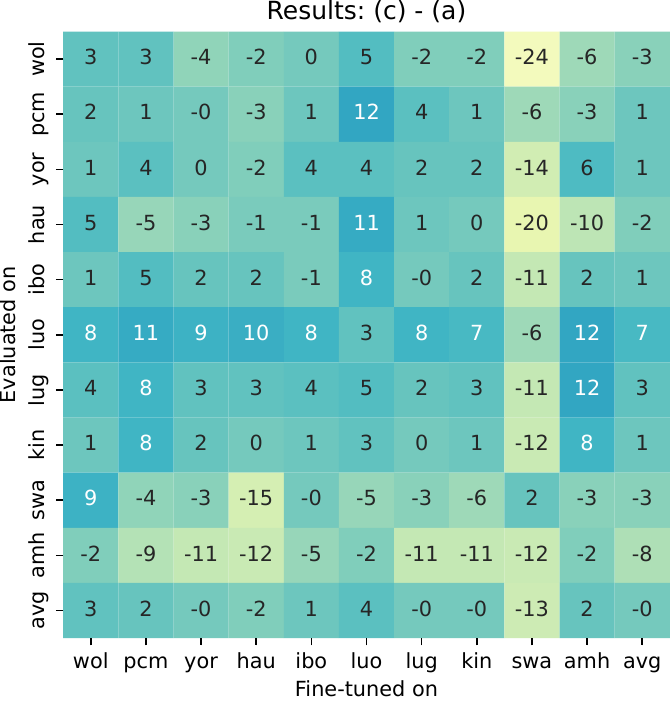}
  \caption{Performance difference between using a Swahili adaptively fine-tuned model and no language-adaptive fine-tuning after subsequent fine-tuning on NER data. Hausa fine-tuning performs much worse when evaluated on Swahili.}
  \label{fig:swa_results_heatmap_diff}
\end{subfigure}
\caption{Heatmaps indicating the average performance over 5 seeds of specific models on specific languages (y-axis) after being fine-tuned on another language's NER data (x-axis). In general, we notice a large standard deviation, indicating that this process is unreliable. The bottom row shows the difference between one technique, and base, i.e. how much improvement this new model gives over using the base model. avg indicates the average transfer performance per row or column, respectively. Note that this calculates the average of the entire row or column excluding the diagonal, to be able to see the overall transfer performance at a glance.}
\label{fig:appdx:allheatmaps}
\end{figure*}

We also consider the above results in slightly more detail by looking at each NER category individually, to see if any perform much better or worse than the others. These results are shown in \cref{fig:all_heatmaps_joint}. We generally see that dates transfer poorly, over most languages, particularly for Luo. This could be caused by the differences in writing dates across these languages. Organisations transfer poorly for Amharic, possibly caused by its different script.

\subsection{International Tokens in Overlap}\label{sec:international_transfer_tokens}
We have a few separate categories of international tokens, described shortly. The full data we use can be found in our source code base.
\begin{description}
  \item[Names] This contains a list of common English names and surnames.
  \item[Places] This list contains continents such as Africa, countries such as Russia, cities such as London and states such as Texas. We additionally have the four cardinal directions (North, South, East, West) and the $10\text{ }000$ cities with the largest population.\footnote{\url{https://public.opendatasoft.com/explore/dataset/geonames-all-cities-with-a-population-1000/download/}}.
  \item[Companies] A list of popular companies and organisations, such as Twitter, Youtube, Boeing, etc. We additionally use a list of the Fortune 1000 companies.
  \item[Numbers/Punctuation] This category contains numbers and punctuation marks.
  \item[General English] General English words, such as the names of the 12 months and the 7 days, words such as ``International'', ``Hospital'', ``Christmas'', etc. This category had the fewest occurrences on average, around 1\% of tokens.
\end{description}
We generally find that places, names and numbers made up most of the overlapping tokens. Punctuation, the names of companies and general English words make up the smallest fraction, with less than 10\% of the tokens.

\subsection{Alternate Overlap Calculations}\label{sec:other_overlap_methods}
While we used just a single method for calculating overlap in the main text, here we show the results when using other, reasonable techniques. Overall, we find that the conclusions are the same, with overlap correlating strongly with transfer performance. In particular, the variations we consider are:
\begin{description}
  \item[Unique Entities] Only count the number of unique overlapping entities between the datasets.
  \item[Only Train]  Consider only the training dataset.
  \item[Source/Target/Sum] When counting the number of times a token overlaps, use the number of occurrences in the source dataset, or the target dataset, or the sum of these two values.
  \item[Normalise] Whether or not to normalise the overlap by dividing by either (1) the total number of entities, (2) the number of entities in the source, or (3) target language, etc.
  \item[Considering ``O''] Whether or not to consider the ``Other'' entities as well when calculating overlap, or just using the named entities.
  \item[Without Labels] Whether to consider two entities overlapping if they have different labels.
\end{description}

Overall, we find that using the number of overlapping tokens as the number of occurrences in the source dataset has the lowest correlation, with $R$ around $0.5$. If we do not consider this approach of calculating overlap, then all correlation coefficients are at least $0.6$, ranging up to $0.7$. All correlations are statistically significant, with $p < 0.05$.

This shows that regardless of the overlap method used, there is a strong correlation between the number of overlapping tokens and the transfer performance in NER. Some of the results for different calculation methods are shown in \cref{fig:appdx:all_data_overlap:2}. In particular, in the bottom row of this figure, we show results similar to those in the main text, but considering only the number of tokens present in the target dataset. We also calculate the fraction of overlapping tokens instead of the absolute number.

\newcommand{\wwww}{0.37}
\begin{figure*} 
  \centering
  \begin{subfigure}[t]{\wwww\linewidth}
    \centering\captionsetup{width=.95\linewidth}
    \resizebox{1\linewidth}{1\linewidth}{\includegraphics[width=1\linewidth]{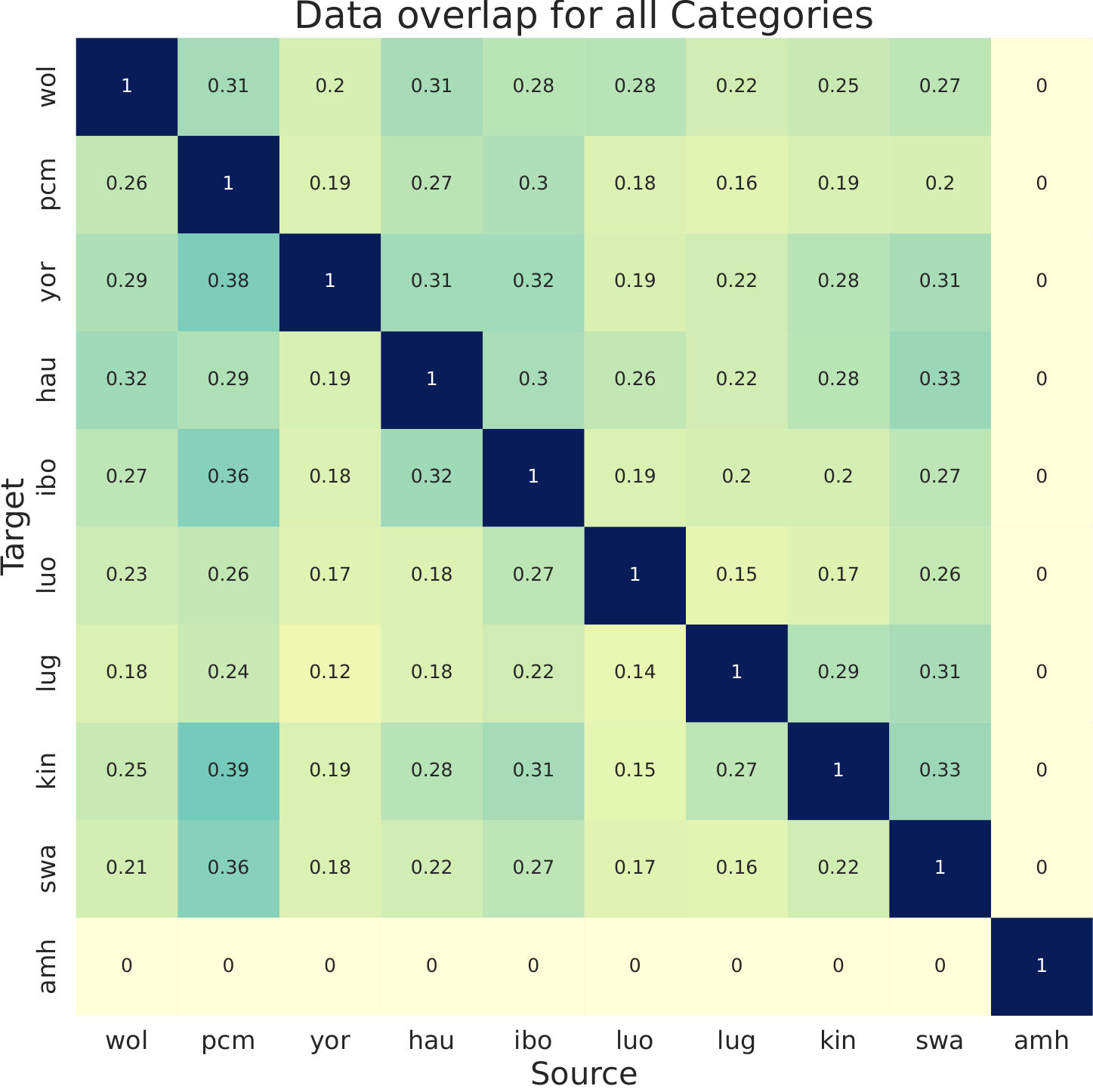}}
    \subcaption*{}
  \end{subfigure} \hspace{0.1\linewidth} 
  \begin{subfigure}[t]{\wwww\linewidth}
    \centering\captionsetup{width=.95\linewidth}
    \resizebox{1\linewidth}{1\linewidth}{\includegraphics[width=1\linewidth]{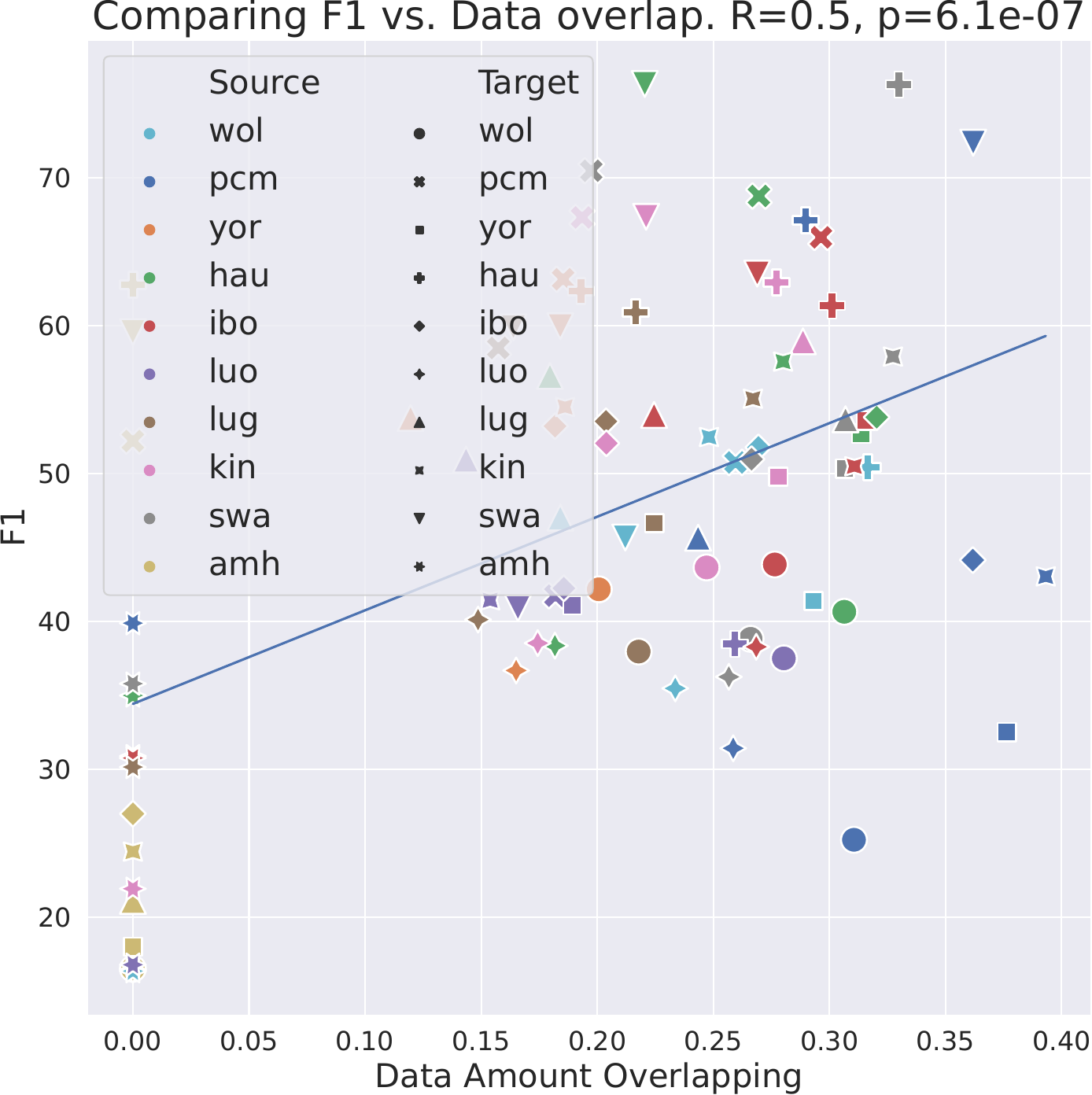
    }}
    \subcaption*{}
  \end{subfigure}

  \begin{subfigure}[t]{\wwww\linewidth}
    \centering\captionsetup{width=.95\linewidth}
    \resizebox{1\linewidth}{1\linewidth}{\includegraphics[width=1\linewidth]{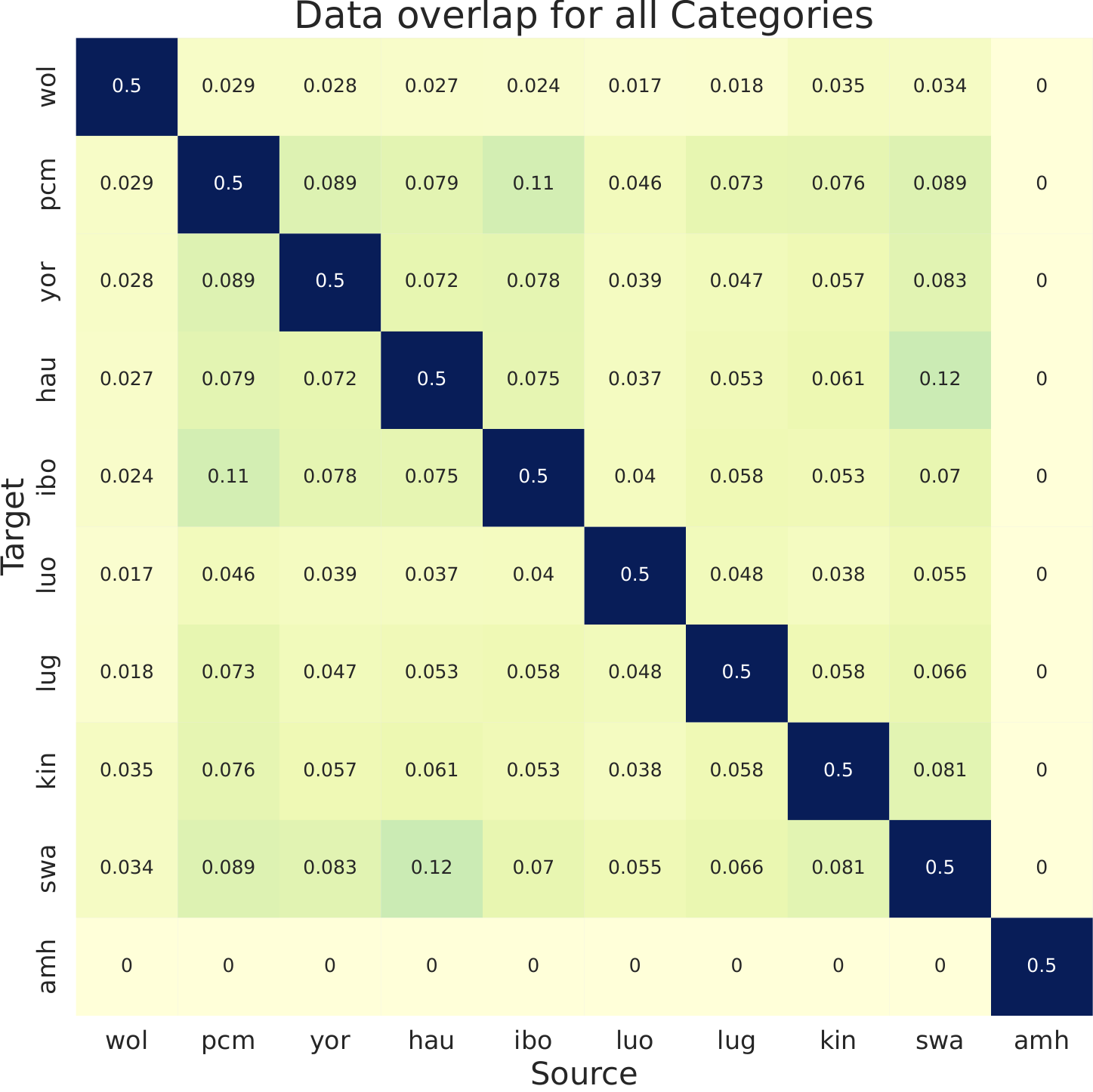}}
    \subcaption*{}
  \end{subfigure} \hspace{0.1\linewidth} 
  \begin{subfigure}[t]{\wwww\linewidth}
    \centering\captionsetup{width=.95\linewidth}
    \resizebox{1\linewidth}{1\linewidth}{\includegraphics[width=1\linewidth]{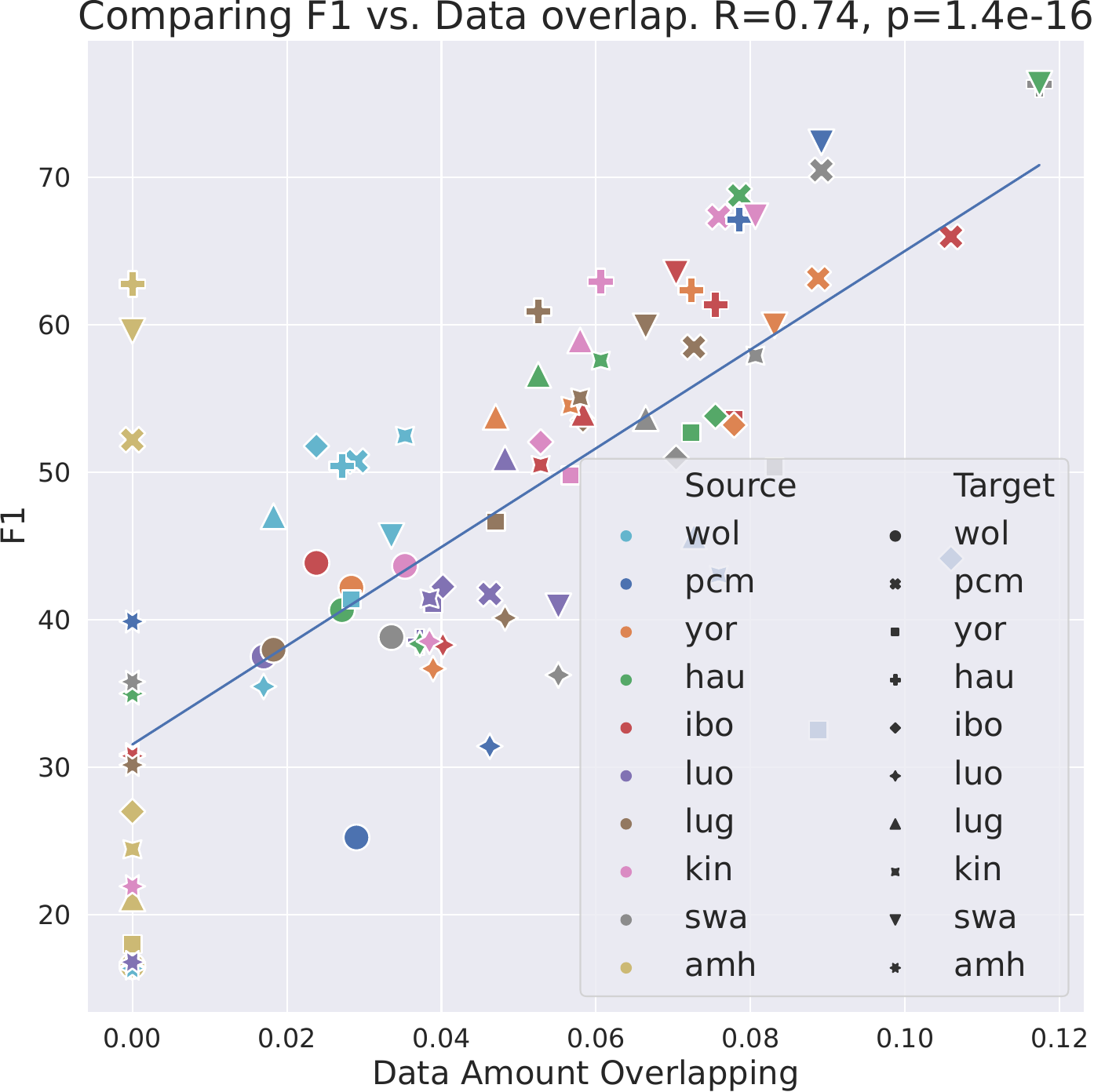}}
    \subcaption*{}
  \end{subfigure}

  \begin{subfigure}[t]{\wwww\linewidth}
    \centering\captionsetup{width=.95\linewidth}
    \resizebox{1\linewidth}{1\linewidth}{\includegraphics[width=1\linewidth]{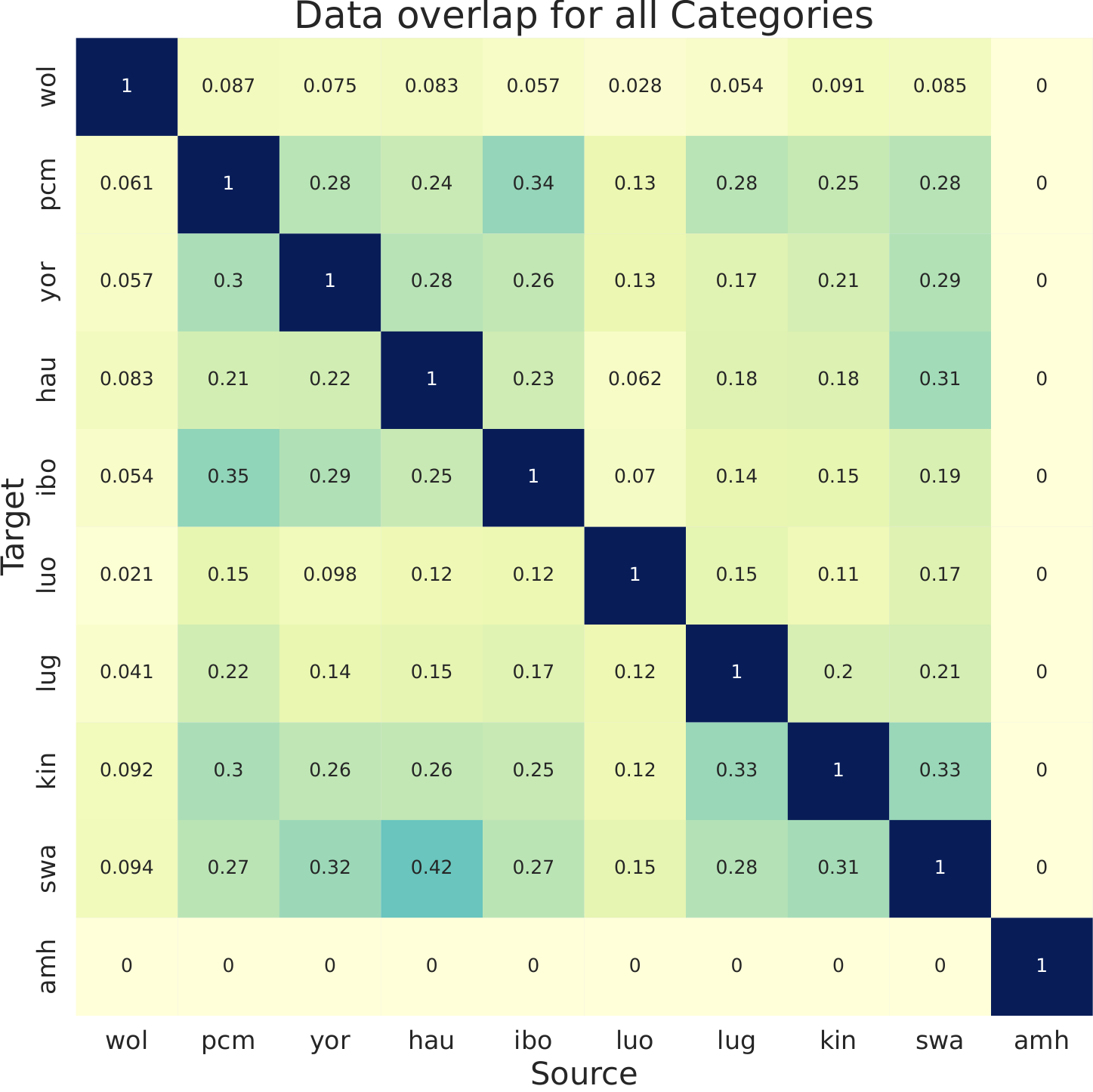}}
  \end{subfigure} \hspace{0.1\linewidth} 
  \begin{subfigure}[t]{\wwww\linewidth}
    \centering\captionsetup{width=.95\linewidth}
    \resizebox{1\linewidth}{1\linewidth}{\includegraphics[width=1\linewidth]{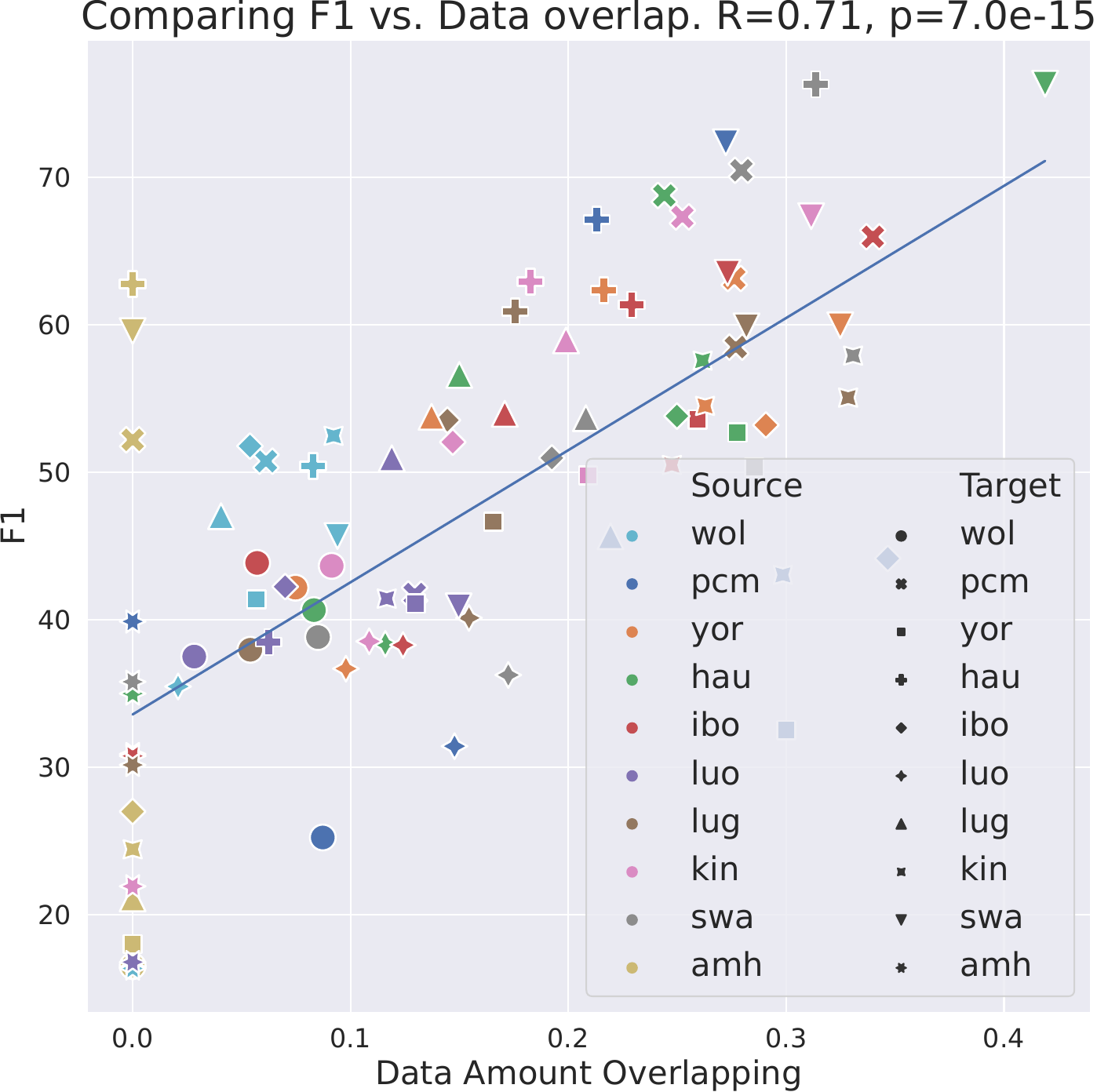}}
  \end{subfigure}
  \caption{Overlap and correlation plots, for the (top) smallest and (middle) largest correlation coefficients, respectively. The bottom row contains the results when calculating overlap as the fraction of overlapping tokens in the target dataset, to contrast against the main text that used the absolute number.
  The top row used just the training dataset, counted overlap with respect to the number of occurrences in the source datasets, without considering labels. The middle row also used all of the unlabelled data, but calculated $\frac{|E_s \cap E_t|}{|E_s| + |E_t|}$ where $E_s$ and $E_t$ are the sets of unique entities for the source and transfer languages respectively~\citep{choosing_transfer_languages}}
  
  \label{fig:appdx:all_data_overlap:2}
  \end{figure*}

\subsection{Splitting Overlap into Local and International}\label{sec:split_overlap_local_international}
\begin{figure*} 
  \centering
  \begin{subfigure}[t]{\wwww\linewidth}
    \centering\captionsetup{width=.95\linewidth}
    \resizebox{1\linewidth}{1\linewidth}{\includegraphics[width=1\linewidth]{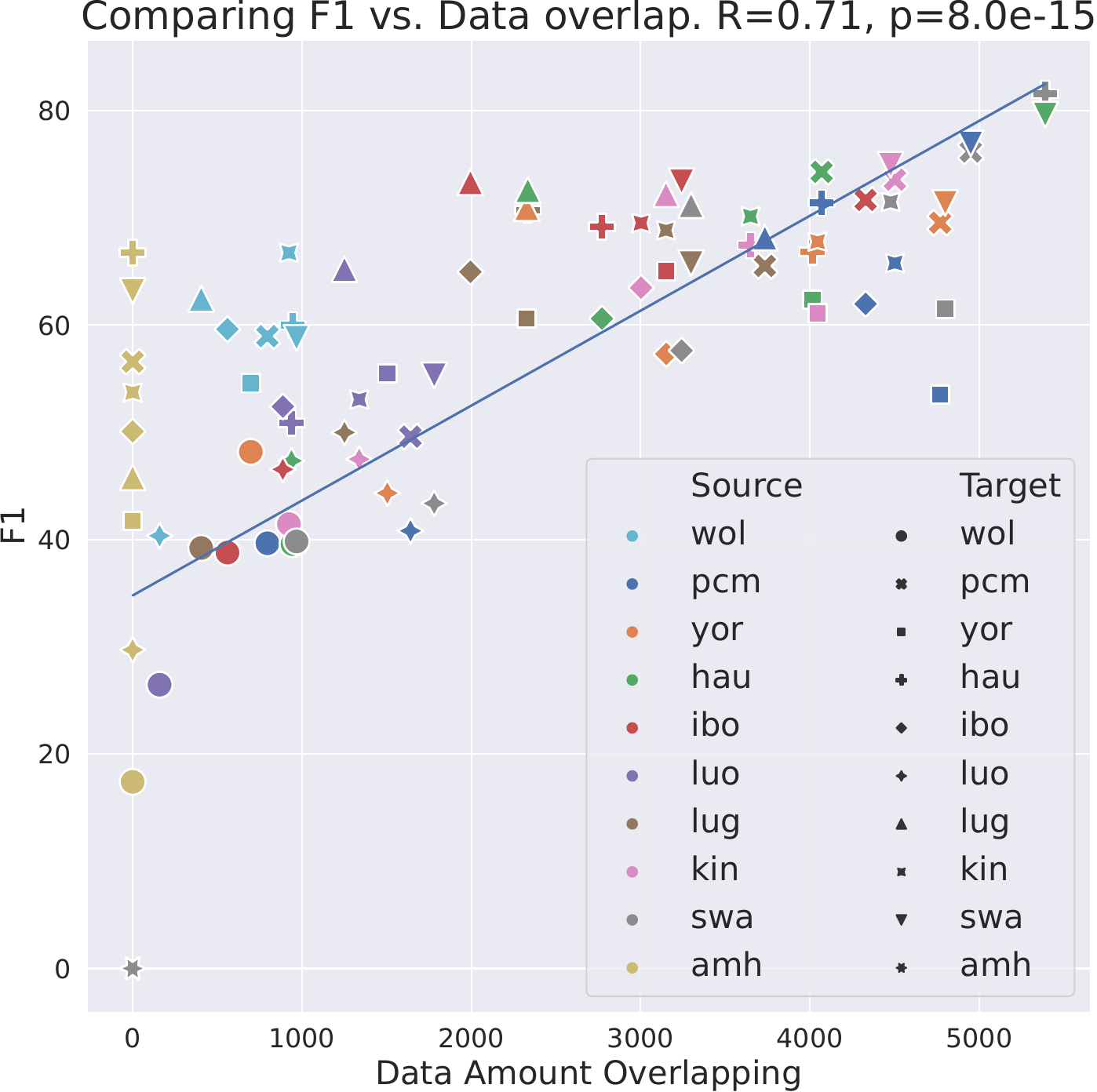}}
    \subcaption{International}
  \end{subfigure} \hspace{0.1\linewidth} 
  \begin{subfigure}[t]{\wwww\linewidth}
    \centering\captionsetup{width=.95\linewidth}
    \resizebox{1\linewidth}{1\linewidth}{\includegraphics[width=1\linewidth]{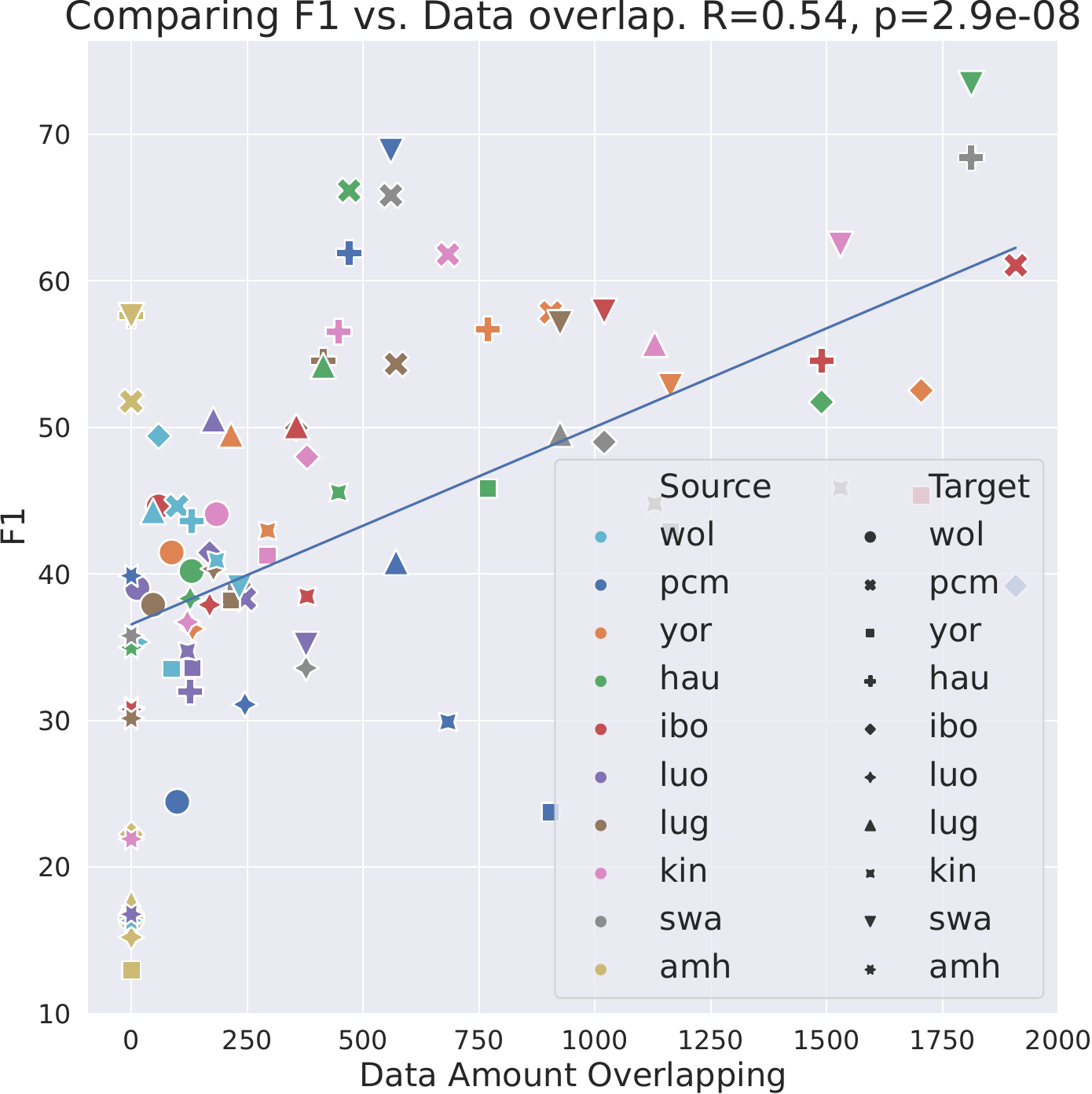
    }}
    \subcaption{Local}
  \end{subfigure}
  \caption{Showing the correlation between overlap and performance when only considering (a) International and (b) Local tokens. Here, both the performance and overlap calculations only took these subsets of tokens into account, for instance, comparing the number of overlapping international tokens with the performance on international tokens.}
  \label{fig:appdx:all_data_overlap:split_local_global}
  \end{figure*}

See \cref{fig:appdx:all_data_overlap:split_local_global} for the overlap results (similar to \cref{fig:all_data_overlap}), split up into international and local tokens. The results here are similar to the ones in the main text (which was averaged over all tokens). The correlation is slightly lower for local tokens, but it is still positive and statically significant.

\subsection{Overlap Correlations without Amharic}\label{sec:appdx:correlation_no_amharic}
\begin{figure}[h] 
  \centering
    \centering\captionsetup{width=.95\linewidth}
    \resizebox{1\linewidth}{1\linewidth}{\includegraphics[width=1\linewidth]{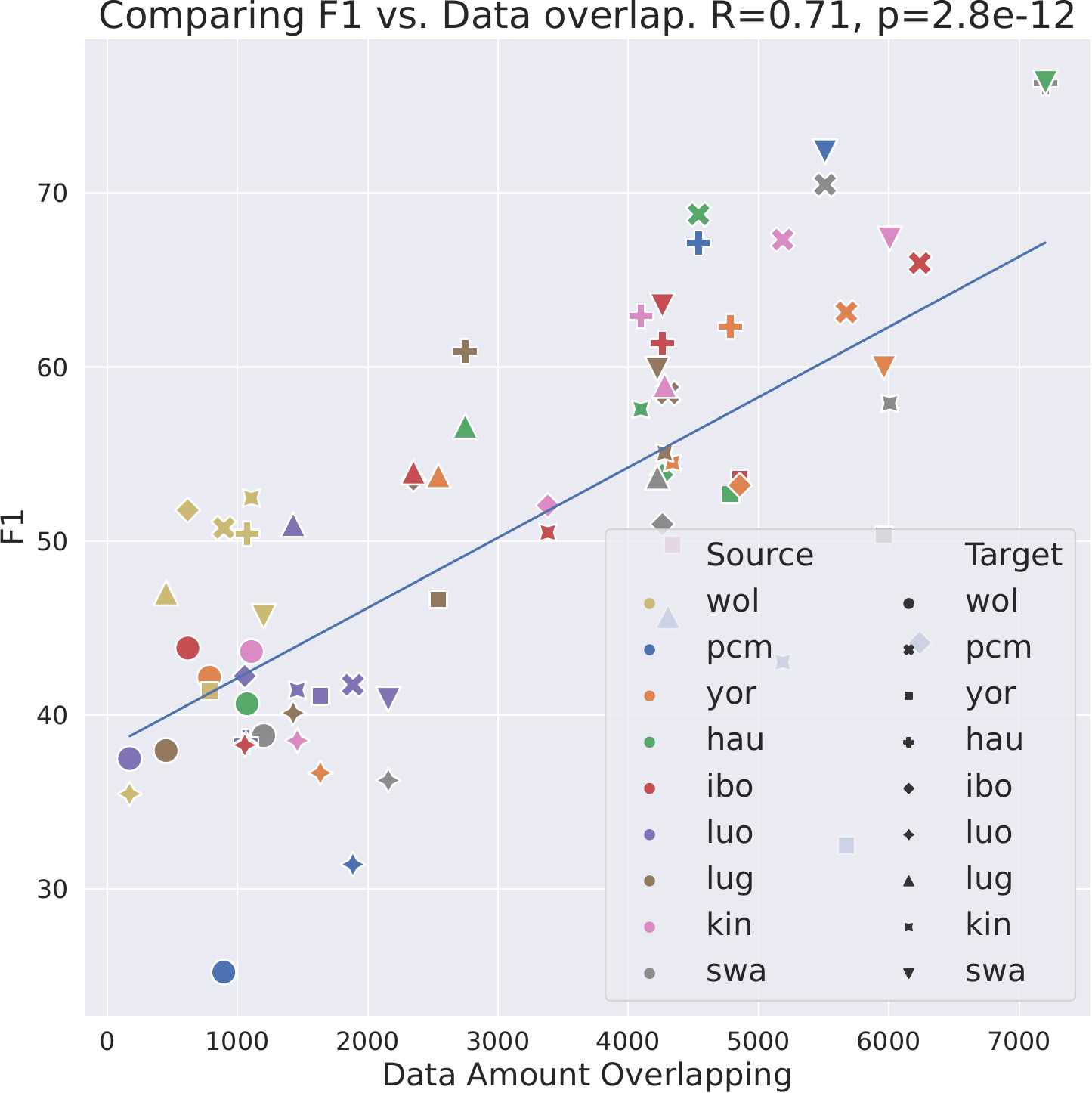}}
  \caption{This shows the correlation between data overlap and performance for Amharic, as it has a different script and may thus be considered an outlier. The results are very similar to \cref{fig:data_corr}.}
  \label{fig:appdx:correlation_no_amharic}
\end{figure}

\cref{fig:appdx:correlation_no_amharic} contains the correlation results when not considering Amharic.

\subsection{Additional Features}\label{sec:other_features}
We additionally consider the other features from \citet{choosing_transfer_languages}. These results are shown in Figures \ref{fig:linetal:subplots_base_Transferovertargetsizeratio}-\ref{fig:linetal:subplots_base_GEOGRAPHIC}. In particular, we consider the following features:
\begin{description}
  \item[Geographic distance] The distance between where the different languages are spoken, based on data from Glottolog~\citep{hammarstrom2018glottolog}.
  \item[Genetic distance] The genealogical distance between the languages based on the Glottolog language tree.
  \item[Inventory distance] The cosine distance between the feature vectors from the PHOIBLE database~\citep{moran2014phoible}.
  \item[Syntactic distance] The cosine distance between the feature vectors that represent the syntactic properties of the languages, from the WALS database~\citep{wals}.
  \item[Phonological distance] The cosine distance between the phonological feature vectors obtained from WALS and Ethnologue databases~\citep{lewis2009ethnologue}.
  \item[Featural distance] The cosine distance between feature vectors consisting of the 5 above features.
  \item[Source language dataset size] The number of sentences in the source language's dataset.
  \item[Source Over Target Size Ratio] The size (in number of sentences) of the source dataset divided by the size of the target dataset.
  \item[Source language number of entities] The number of named entities in the source language's dataset.
  \item[Source Over Target entity Ratio] The number of entities in the source dataset divided by the number of entities in the target dataset.
\end{description}

Overall, we find that data overlap has the highest correlation with transfer performance, with many other features not having a statistically significant correlation or a very small positive or negative correlation.

\subsection{Combining Datasets}\label{sec:combiningdatasets}
      As an additional experiment, we train models on a combination of datasets, to see if this has any effect. The two options we consider here are: (1) training on the concatenation of all of the datasets; and (2) training on the concatenation, excluding the target language. We consider (1), as it involves training a single model, and we would like to investigate how well this model performs across all languages. For (2), we measure the effect of transferring from the nine other languages, as opposed to only the single-language transfer we have considered in the rest of this paper.
      These results are shown in \cref{tab:combination_per}. Training on the target language, or on the concatenation of all languages performs quite well. The latter option also has the advantage of only being one model, whereas we need one model per language if we fine-tune only on data from one language.
      Fine-tuning the base model on the best transfer language performs worse than training on all of the datasets, excluding the target language. Finally, using a LAFT model for the target language and fine-tuning on all datasets except the target performs much better, and is the best transfer option we have considered.
  
      \begin{table}[H]
        \centerfloat
        \caption{Here we show the results when training on a combination of datasets, compared to training on the best transfer language, or the target language itself. $cat - \{X\}$ indicates that the model trained on a combination of all of the datasets excluding the target language.}
        \label{tab:combination_per}
        \begin{adjustbox}{width=1\linewidth}
          \begin{tabular}{lrrrrrrrrrrr}
\toprule
{} &  wol &  pcm &  yor &  hau &  ibo &  luo &  lug &  kin &  swa &  amh &  avg \\
\midrule
$\text{base} \to X$                        &   64 &   87 &   78 &   89 &   85 &   74 &   80 &   74 &   88 &   71 &   79 \\
$\text{base} \to \text{X} \to {X}$         &   67 &   87 &   83 &   92 &   88 &   76 &   85 &   78 &   90 &   77 &   82 \\
$\text{base} \to cat$                      &   65 &   89 &   81 &   91 &   86 &   77 &   81 &   75 &   87 &   71 &   80 \\
$\text{base} \to \text{swa} \to cat$       &   66 &   89 &   80 &   91 &   85 &   80 &   81 &   76 &   89 &   69 &   81 \\
$\text{base} \to \text{X} \to cat - \{X\}$ &   57 &   80 &   68 &   78 &   76 &   45 &   73 &   68 &   74 &   49 &   67 \\
$\text{base} \to cat - \{X\}$              &   43 &   78 &   60 &   74 &   60 &   44 &   63 &   59 &   75 &   33 &   59 \\
$\text{base} \to best$                     &   44 &   70 &   54 &   76 &   54 &   40 &   59 &   58 &   76 &   40 &   57 \\
\bottomrule
\end{tabular}

          \end{adjustbox}
      \end{table}

\subsection{Representations}\label{sec:embeddings}
This additional experiment follows prior work by \citet{zero_shot_transfer} by investigating the contextual word embeddings from the different models, specifically looking into how these embeddings change as we perform different fine-tuning operations.
We take the last 4 layers from the language model (i.e. not the dense final layer) and use the sum of these hidden states to obtain a word vector (of size 768). We use the sentences from the dataset, and only extract the 4 different NER categories for computational reasons. We compute the mean vector per category, which we use in the following.
To visualise the data, we show the results after performing PCA. 
\subsubsection{Variability}
We found a large amount of variability when fine-tuning the models on different random seeds (see Figure 1 in the main text), so we next investigate the effect of different initialisations on the embeddings.

\cref{fig:all_embed} shows the results for a few languages pairs, and immediately we can see that \cref{fig:embed_hau_then_hau} has clusters corresponding to the different categories, even when using different seeds. 
\cref{fig:embed_kin_then_hau,fig:embed_wol_then_swa,fig:embed_wol_then_hau} on the other hand cluster more toward seeds, so the categories differ when using different seeds. This could indicate that the Swahili model is more consistent and robust to random initialisations, and learns roughly the same embeddings for each seed. On the other hand, when fine-tuning from Kinyarwanda, Luo or Wolof, there is no clear clustering of categories (despite a relatively large amount of data overlap between Kinyarwanda and Hausa), suggesting that these models cannot distinguish Hausa categories very well (possibly substantiated by the poorer results shown in the main text).

Now, the above analysis is somewhat impacted by the final linear layers in the models -- it is entirely possible that two models that have different embeddings also have different final layers and end up classifying examples exactly the same. We can, however, still use these experiments to extract some qualitative information about the embeddings of different languages. Furthermore, \cref{fig:embed_swa_swa_v2,fig:embed_hau_hau_v2} -- in which the language being investigated is the same as what the models trained on -- contain results where the clustering is predominantly towards categories, bolstering the validity of this approach.

\subsubsection{Different Languages and Models}
Here we consider the same model and analyse the differences in embeddings from different languages, and how this evolves.
For example, in \cref{fig:pcm_diff_lang} we see that for Nigerian Pidgin (which transferred well previously), the predominant clusters are again categories and not languages.

We next examine different models on the same language, specifically looking at what happens to these embeddings when a model is further fine-tuned. \cref{fig:hau_diff_model} shows that performing fine-tuning on models does affect the embeddings quite significantly, although there does still seem to be a similar relative positioning between the categories - almost as if in PCA, one principal component was the model used, and another was the category.

\subsubsection{Transfer when fine-tuning on Amharic}
In the main text, we observed that Amharic transferred quite well to Hausa, Swahili and Nigerian Pidgin.
We now plot the embeddings of different languages, using \textit{base} fine-tuned on Amharic in \cref{fig:appdx:all_amh}. In the top row, we have Hausa, Swahili -- languages that Amharic was jointly pre-trained with -- and Nigerian Pidgin, which is similar to English. In the bottom row, we have three other languages, not contained in the pre-training dataset. Clearly, the top row is clustered significantly more towards categories -- indicating that the model manages to transfer knowledge from Amharic to these other languages. The bottom row demonstrates a clear clustering around the random seed -- indicating no real information is transferred.
\subsubsection{Summary}
In summary, plotting the embeddings can shed some light on the representations learned by the model which, in many cases, provides some explanation for the results we obtained. Examining the embeddings can shed some light on this. 

\newcommand{\ww}{0.33\linewidth}
\begin{figure*}
  \begin{subfigure}[t]{\ww}
    \centering\captionsetup{width=.9\linewidth}
    \includegraphics[width=1\linewidth]{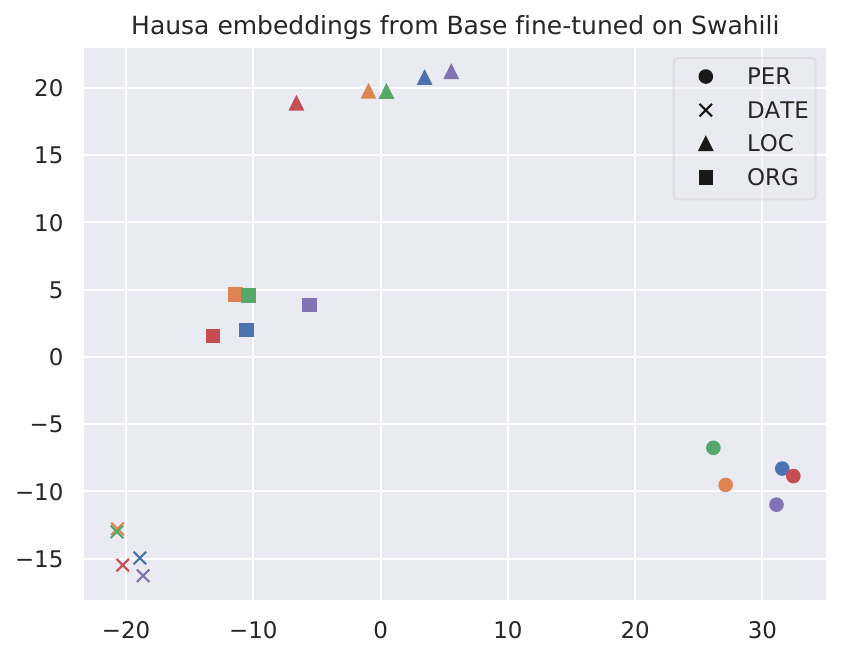}
    \subcaption{Clear category clusters.}
    \label{fig:embed_hau_then_hau}
  \end{subfigure}\hfill
  \begin{subfigure}[t]{\ww}
    \centering\captionsetup{width=.9\linewidth}
    \includegraphics[width=1\linewidth]{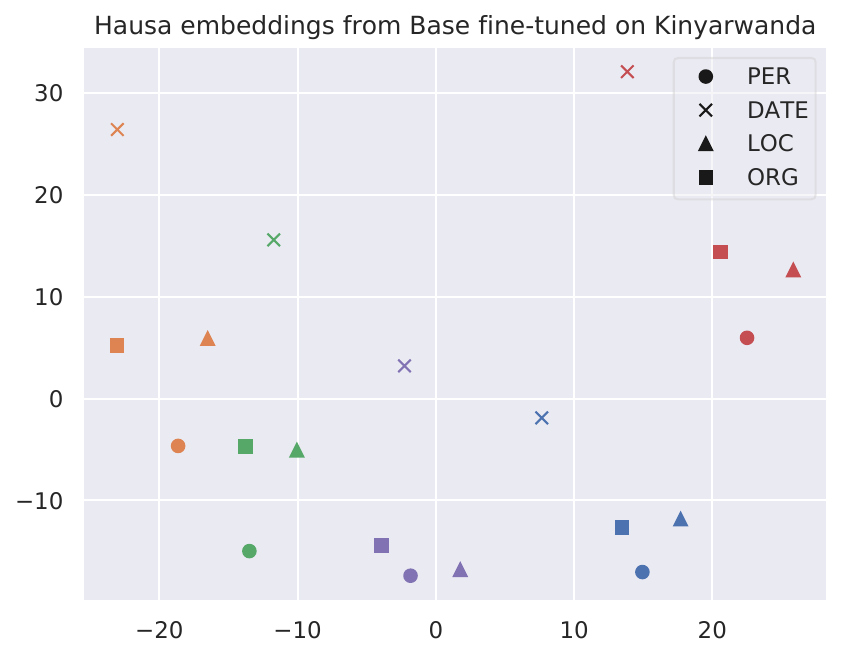}
    \subcaption{Less clear clustering, but largely revolving around colours.}
    \label{fig:embed_kin_then_hau}
  \end{subfigure}\hfill
  \begin{subfigure}[t]{\ww}
    \centering\captionsetup{width=.9\linewidth}
    \includegraphics[width=1\linewidth]{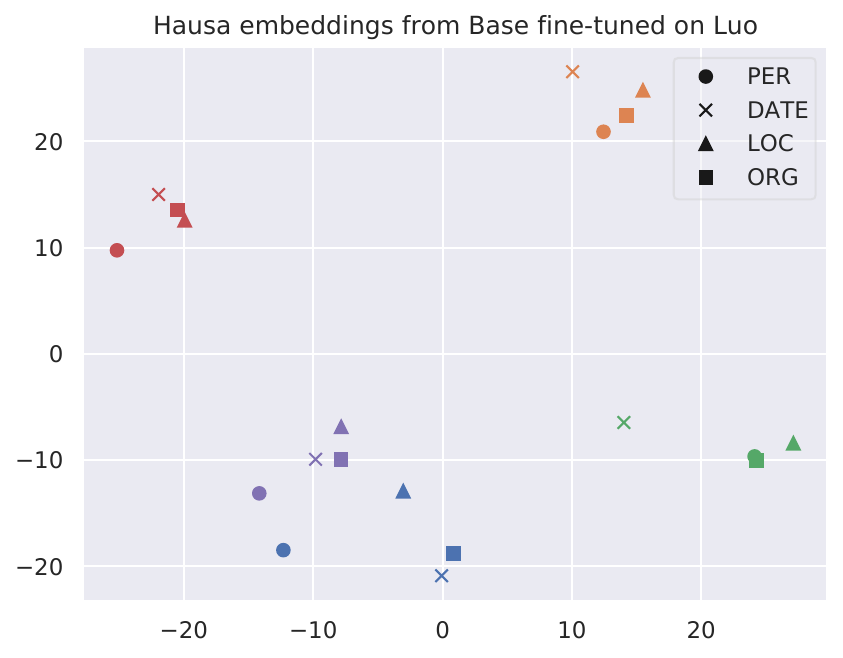}
    \subcaption{Clear colour clusters.}
    \label{fig:embed_wol_then_swa}
  \end{subfigure}\hfill

  \begin{subfigure}[t]{\ww}
    \centering\captionsetup{width=.9\linewidth}
    \includegraphics[width=1\linewidth]{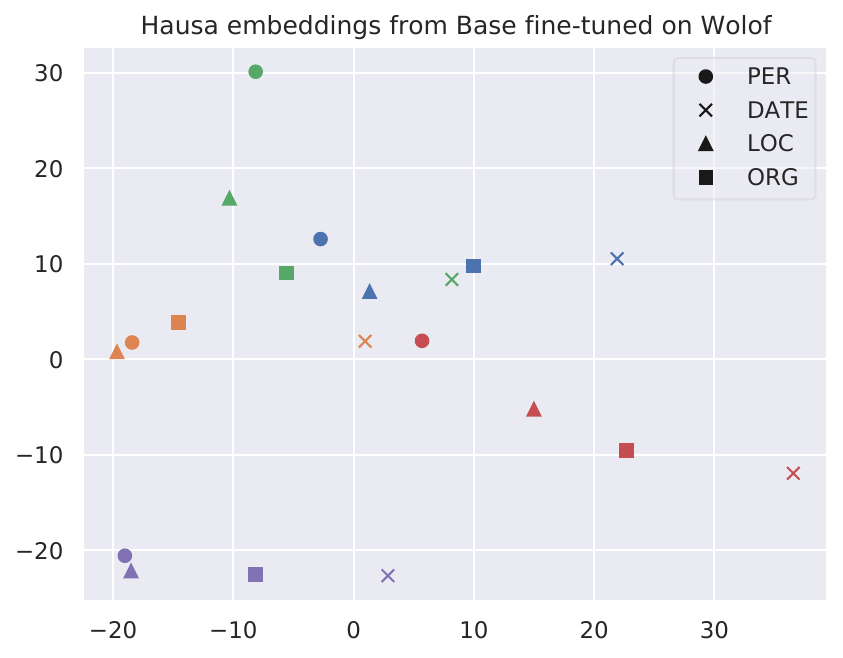}
    \subcaption{Clear colour clusters.}
    \label{fig:embed_wol_then_hau}
  \end{subfigure}
  \begin{subfigure}[t]{\ww}
    \centering\captionsetup{width=.9\linewidth}
    \includegraphics[width=1\linewidth]{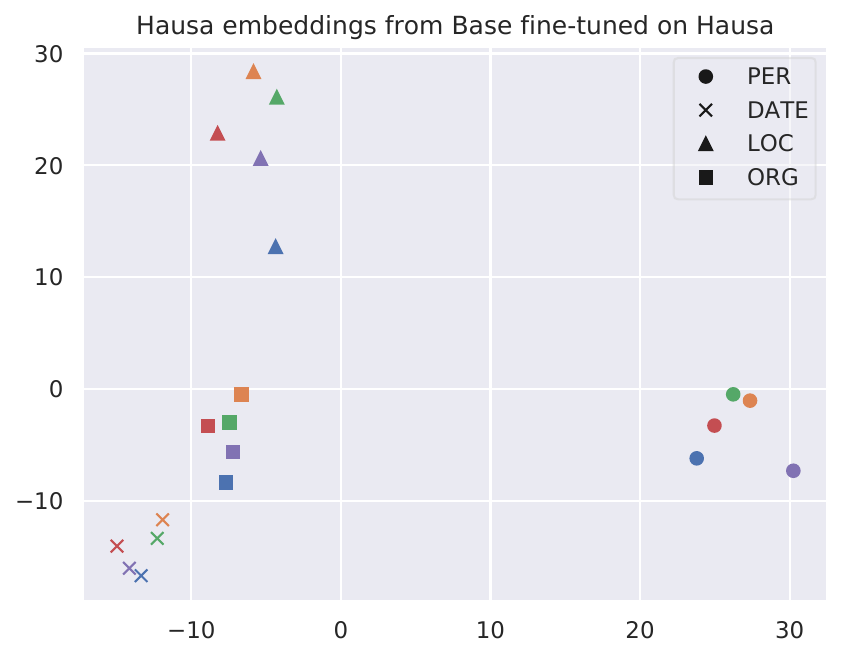}
    \subcaption{Hausa embeddings from a model that was fine-tuned on Hausa -- with clear category clusters.}
    \label{fig:embed_hau_hau_v2}
  \end{subfigure}\hfill
  \begin{subfigure}[t]{\ww}
    \centering\captionsetup{width=.9\linewidth}
    \includegraphics[width=1\linewidth]{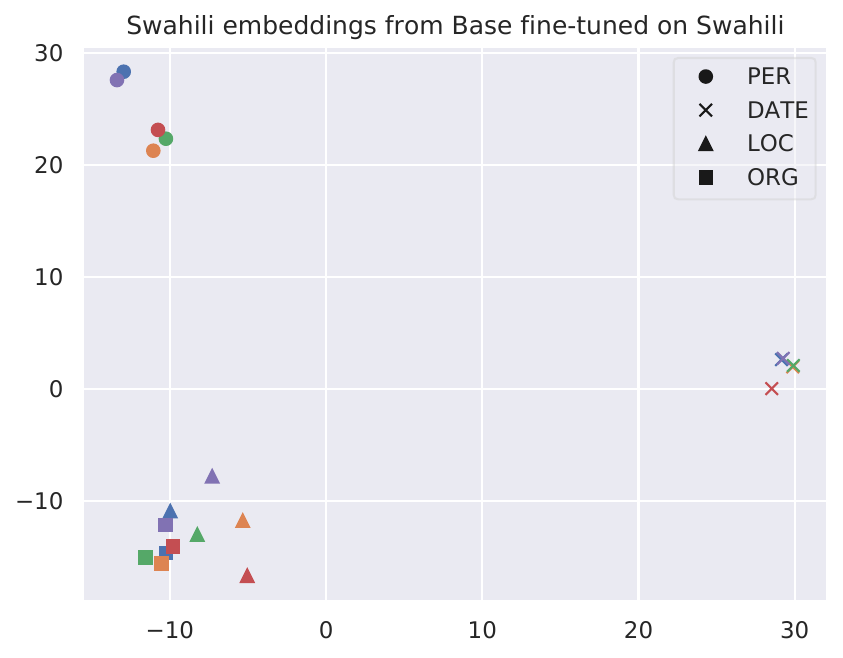}
    \subcaption{Swahili embeddings from a Swahili fine-tuned model -- with clear category clusters, although LOC and organisations are grouped close together.}
    \label{fig:embed_swa_swa_v2}
  \end{subfigure}\hfill

  \caption{Scatter plots of embeddings from different models, languages and categories. The shapes indicate different categories, whereas the colours indicate different starting points, i.e. seeds.}
  \label{fig:all_embed}
\end{figure*}

\begin{figure*} 
  \begin{subfigure}[t]{0.48\linewidth}
    \centering\captionsetup{width=.8\linewidth}
    \includegraphics[width=1\linewidth]{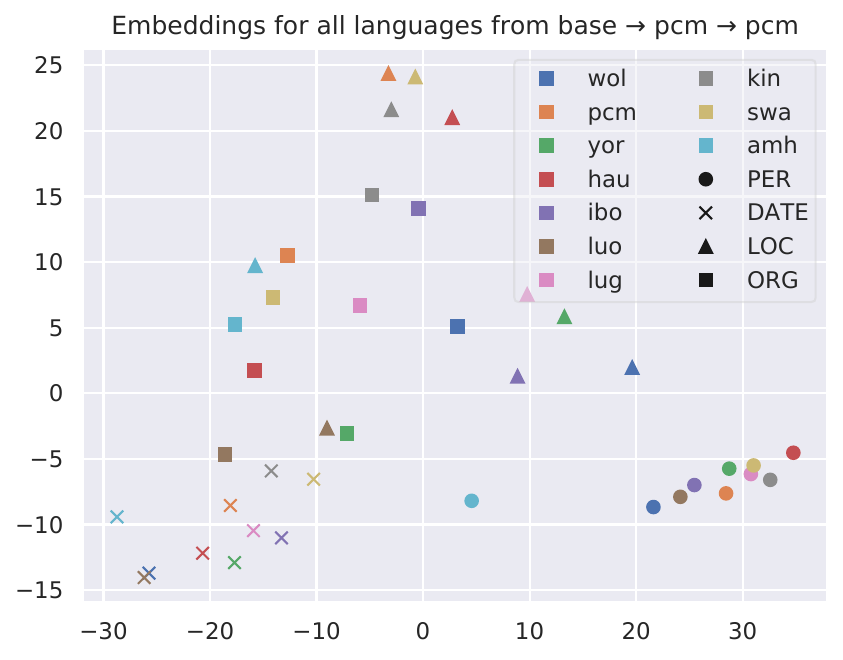}
    \subcaption{Embeddings for different languages from a base~$\to$~pcm~$\to$~pcm model. This clustering is quite similar across different seeds.}
    \label{fig:pcm_diff_lang}
  \end{subfigure}\hfill
  \begin{subfigure}[t]{0.48\linewidth}
    \centering\captionsetup{width=.8\linewidth}
      \includegraphics[width=1\linewidth]{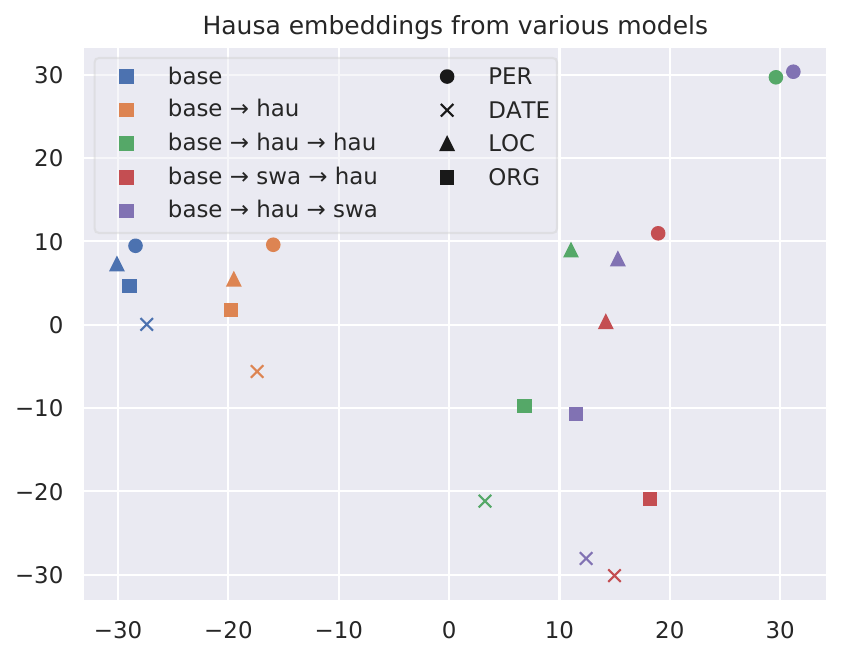}
      \subcaption{Hausa embeddings from different models. }
      \label{fig:hau_diff_model}
    \end{subfigure}
    \caption{Embeddings of (a) multiple languages with one model and (b) Hausa embeddings from different models after performing PCA. }
\end{figure*}

\begin{figure*}
  \begin{subfigure}[t]{\ww}
    \centering\captionsetup{width=.9\linewidth}
    \includegraphics[width=1\linewidth]{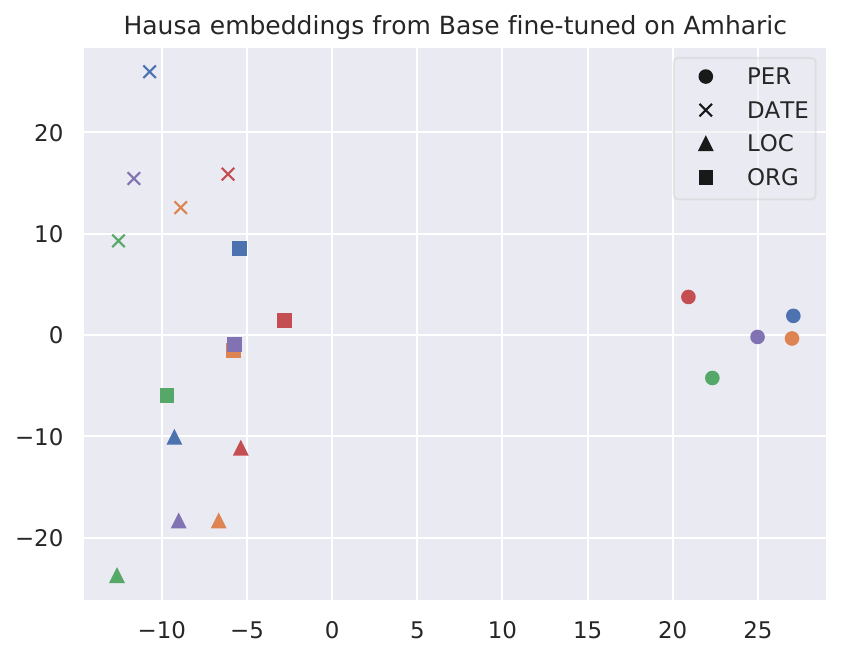}
  \end{subfigure}\hfill
  \begin{subfigure}[t]{\ww}
    \centering\captionsetup{width=.9\linewidth}
    \includegraphics[width=1\linewidth]{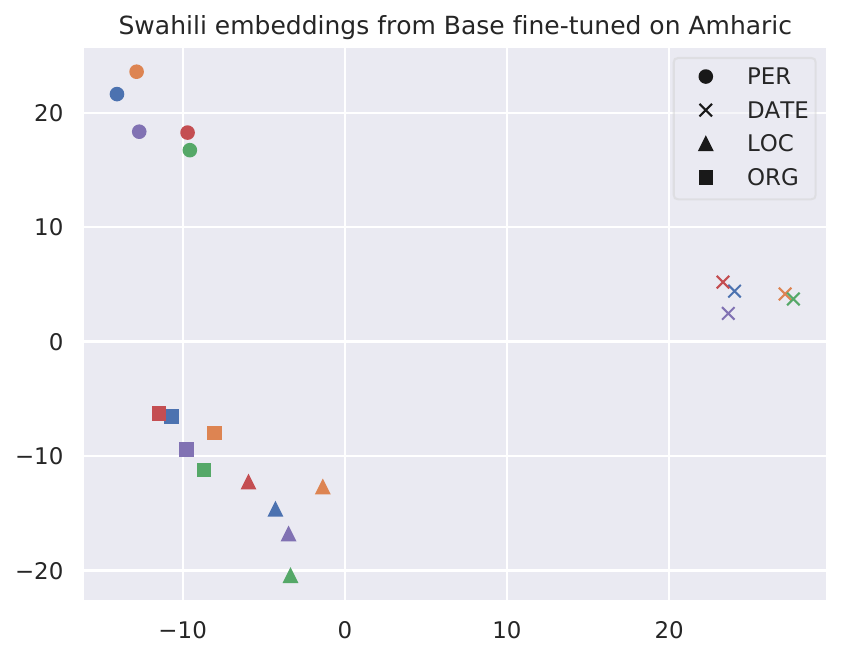}
  \end{subfigure}\hfill
  \begin{subfigure}[t]{\ww}
    \centering\captionsetup{width=.9\linewidth}
    \includegraphics[width=1\linewidth]{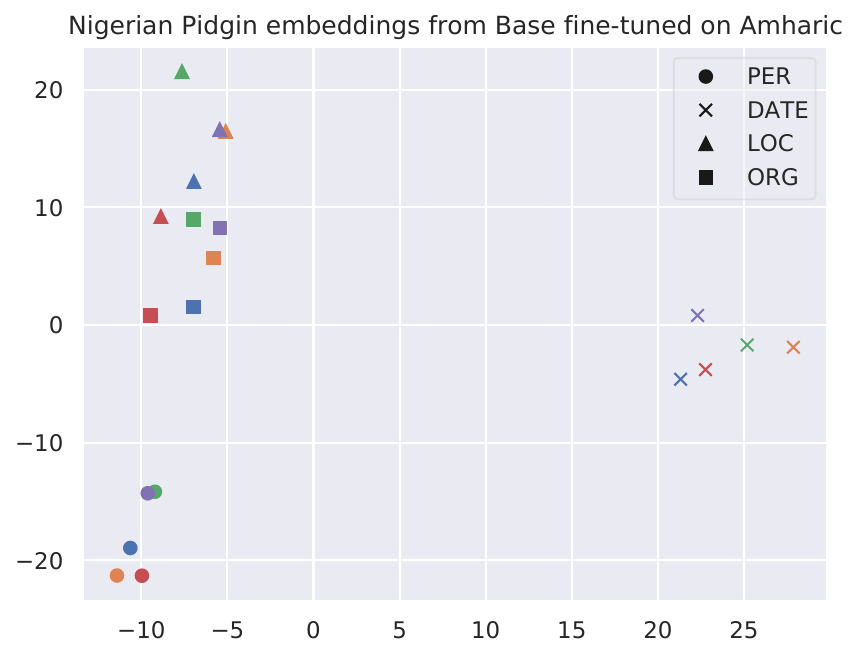}
  \end{subfigure}\hfill

  \begin{subfigure}[t]{\ww}
    \centering\captionsetup{width=.9\linewidth}
    \includegraphics[width=1\linewidth]{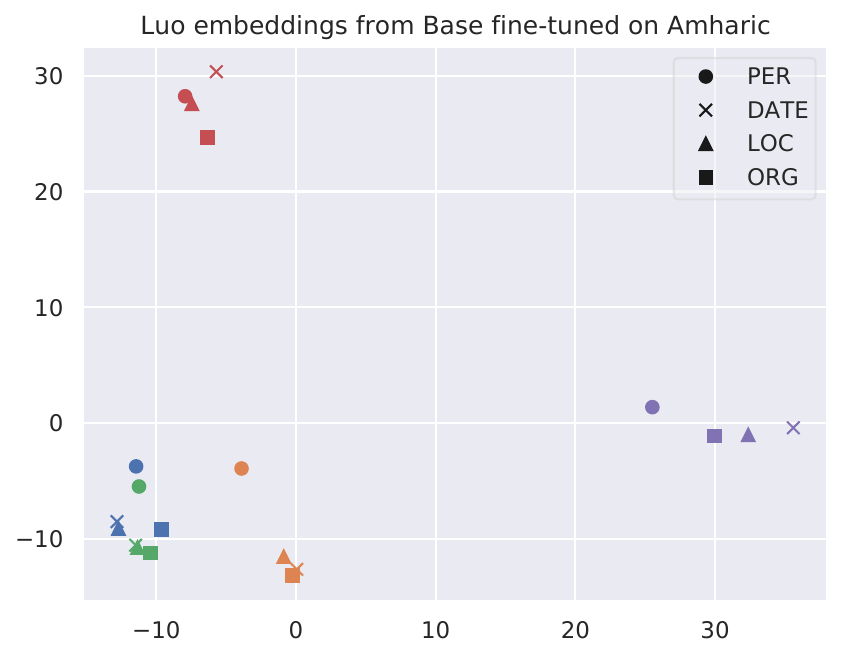}
  \end{subfigure}\hfill
  \begin{subfigure}[t]{\ww}
    \centering\captionsetup{width=.9\linewidth}
    \includegraphics[width=1\linewidth]{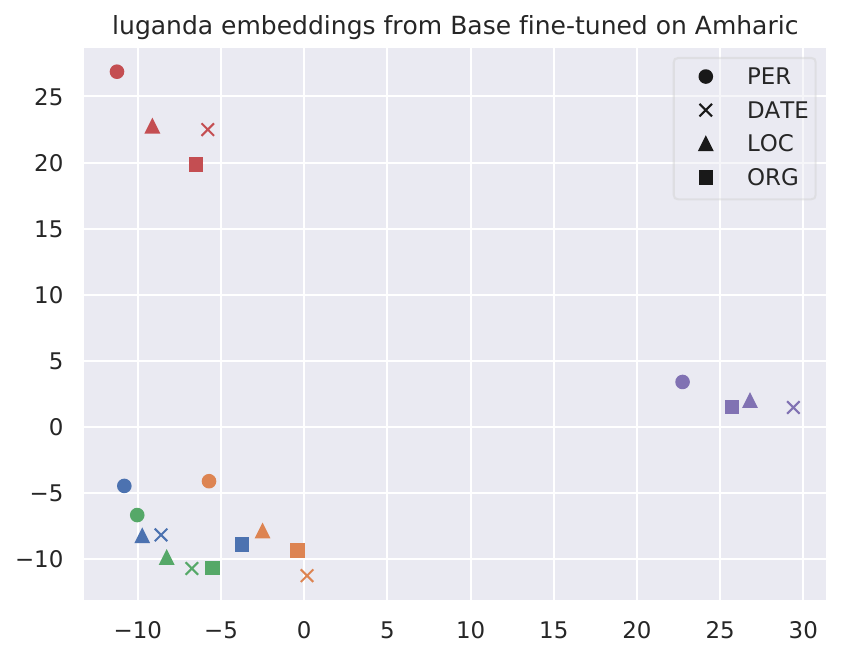}
  \end{subfigure}\hfill
  \begin{subfigure}[t]{\ww}
    \centering\captionsetup{width=.9\linewidth}
    \includegraphics[width=1\linewidth]{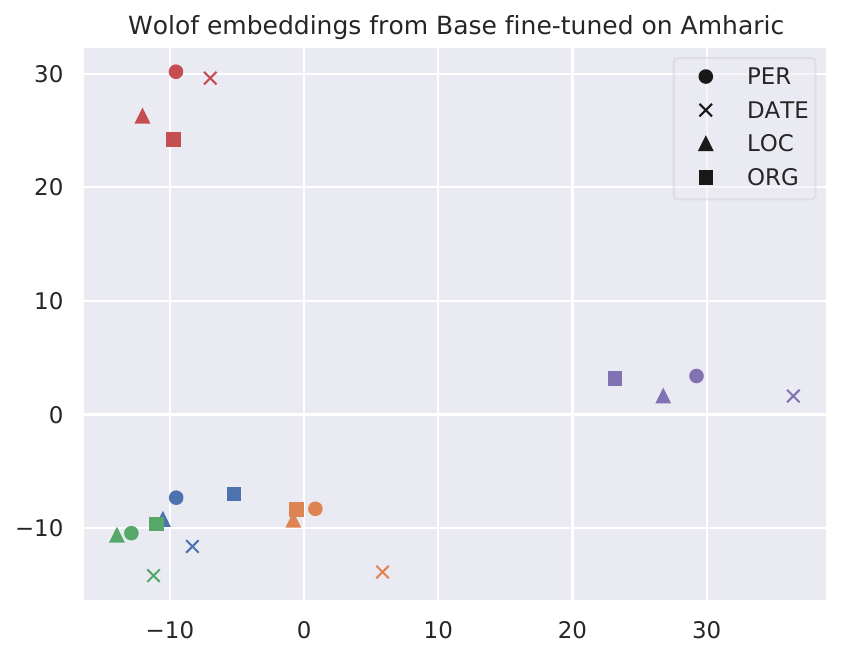}
  \end{subfigure}\hfill
  \caption{Showing embeddings of various languages, obtained from $\text{base} \to \text{amh}$.}\label{fig:appdx:all_amh}
\end{figure*}

\newcommand{\testw}{1}

\begin{figure*}
  \begin{minipage}[t]{1\linewidth}
  \centering\captionsetup{width=.95\linewidth}
  \includegraphics[width=0.9\linewidth]{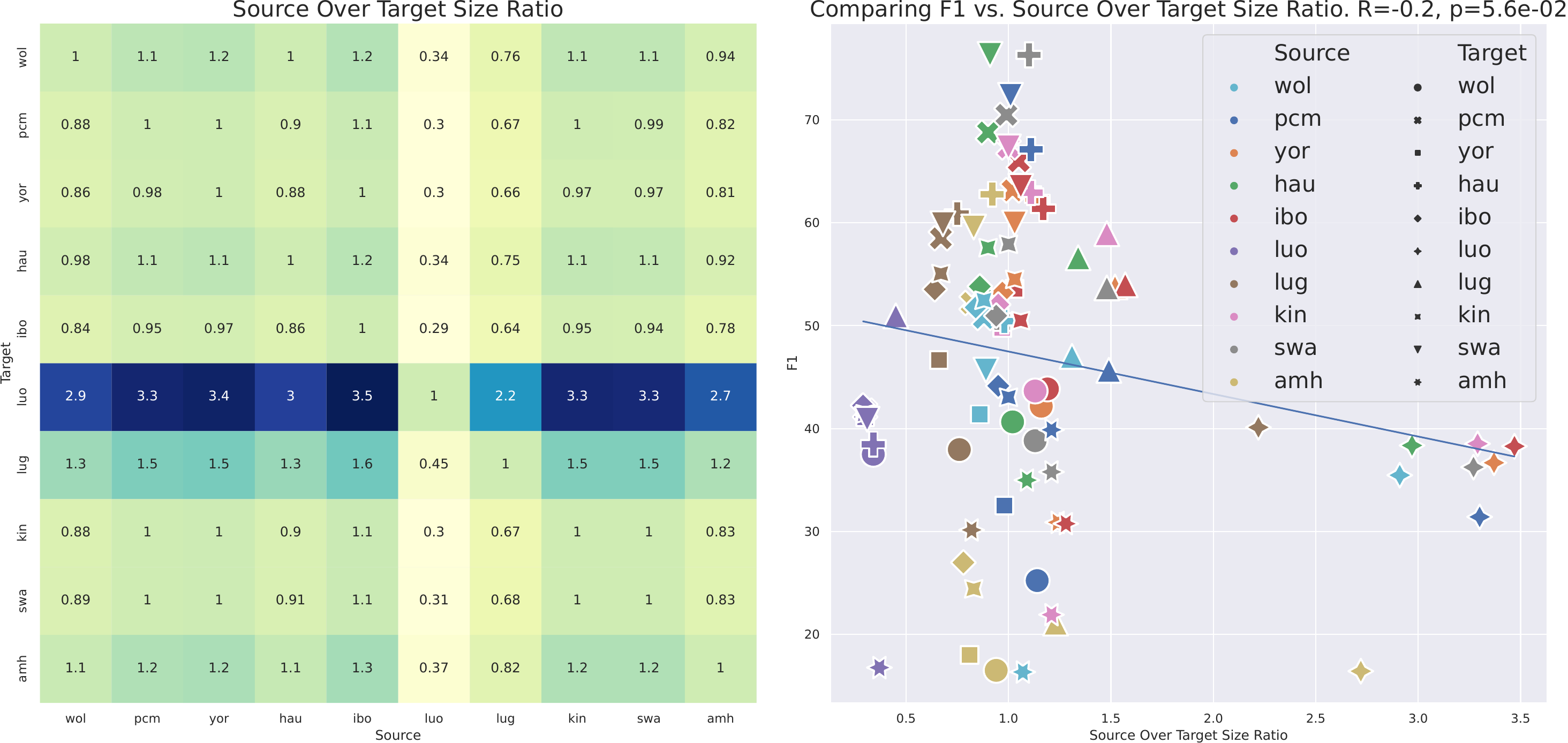}
  \caption{Source over Target Size Ratio}\label{fig:linetal:subplots_base_Transferovertargetsizeratio}

  \centering\captionsetup{width=.95\linewidth}
  \includegraphics[width=0.9\linewidth]{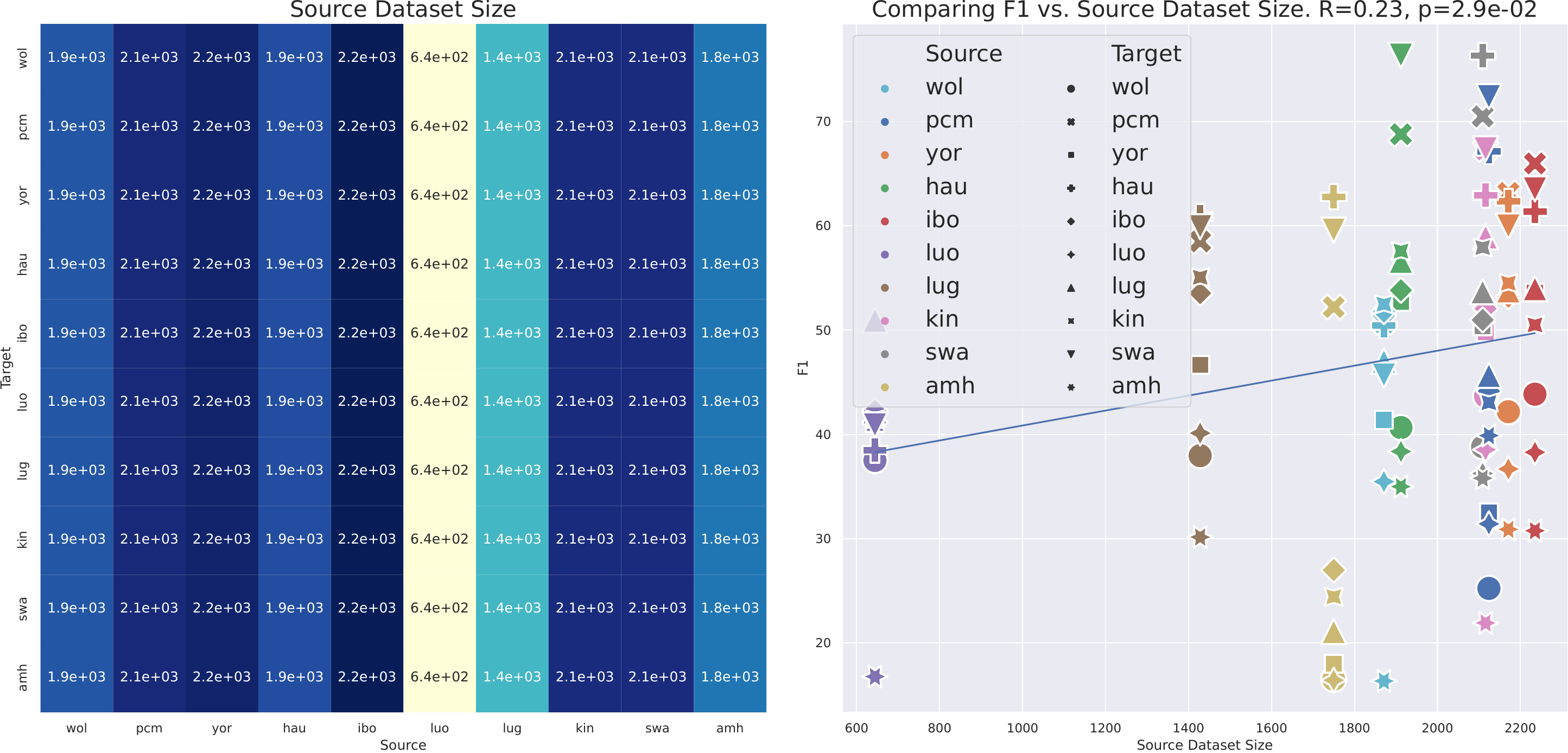}
  \caption{Source Language Dataset Size}\label{fig:linetal:subplots_base_Transferlangdatasetsize}
  \end{minipage}
\end{figure*}

\begin{figure*}
  \begin{minipage}[t]{1\linewidth}
  \centering\captionsetup{width=.95\linewidth}
  \includegraphics[width=0.9\linewidth]{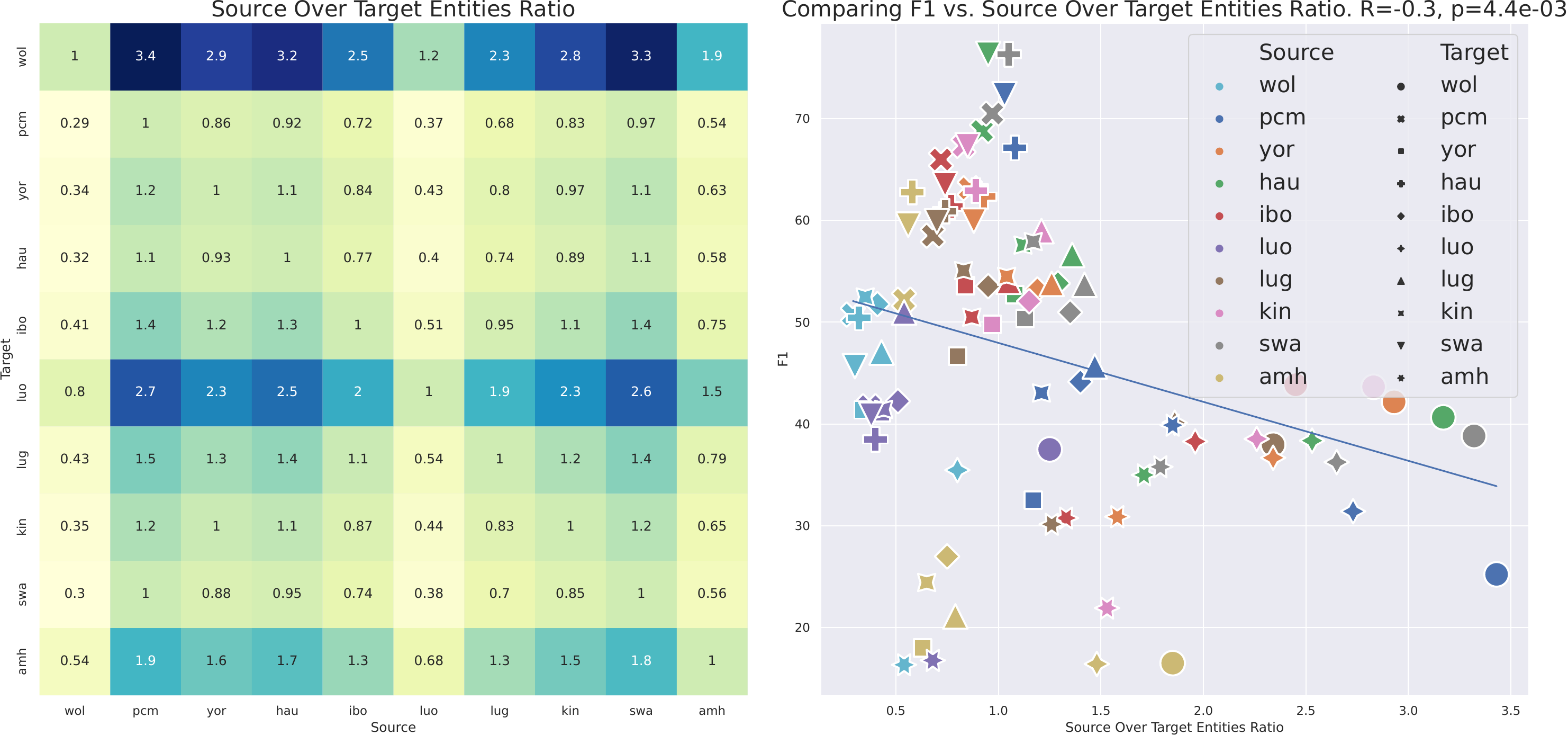}
  \caption{Source over Target Entities Ratio}

  \centering\captionsetup{width=.95\linewidth}
  \includegraphics[width=0.9\linewidth]{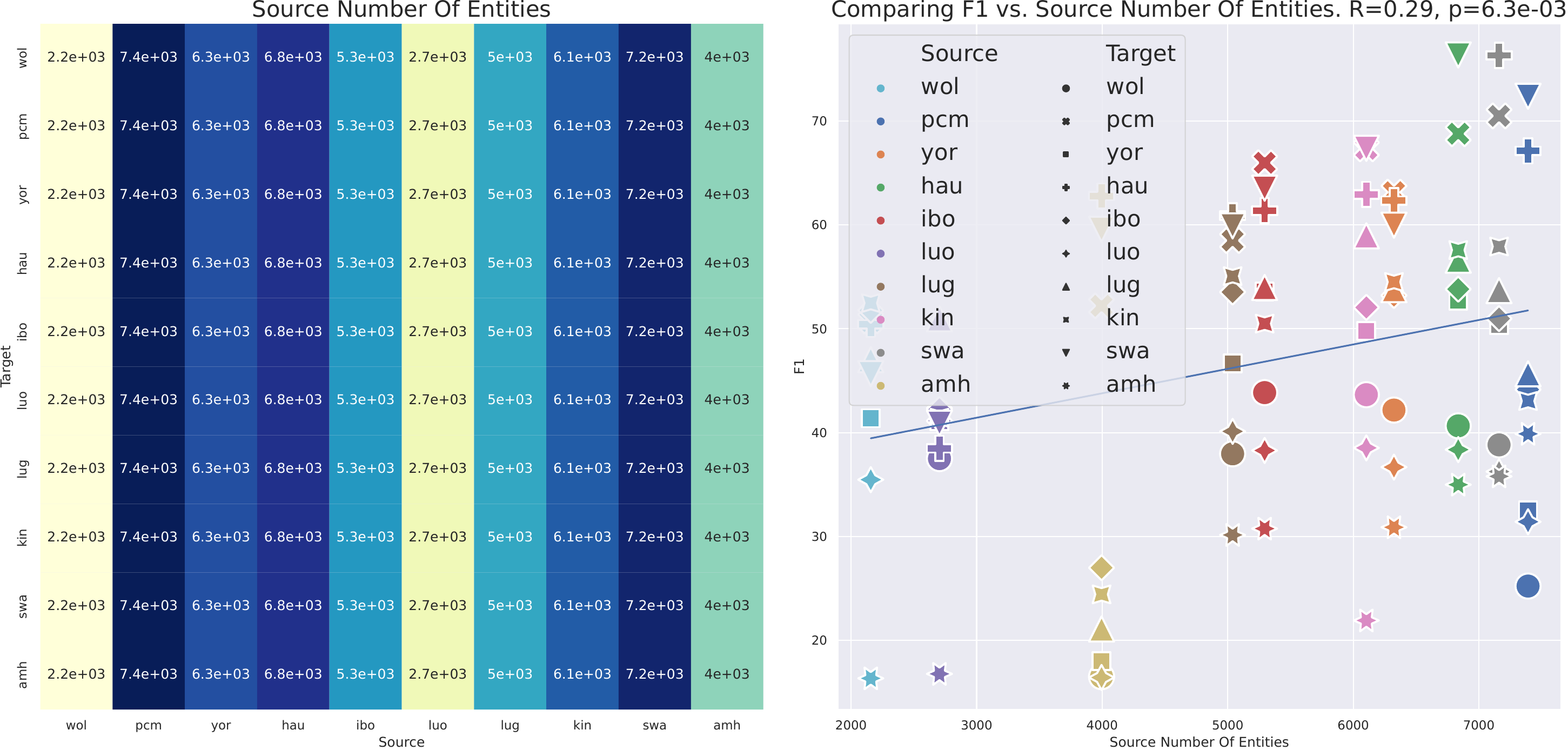}
  \caption{Number of entities in the source language's dataset}
  \end{minipage}
\end{figure*}

\begin{figure*}
  \centering
  \begin{minipage}[t]{1\textwidth}
  \centering\captionsetup{width=.95\linewidth}
  \includegraphics[width=0.9\linewidth]{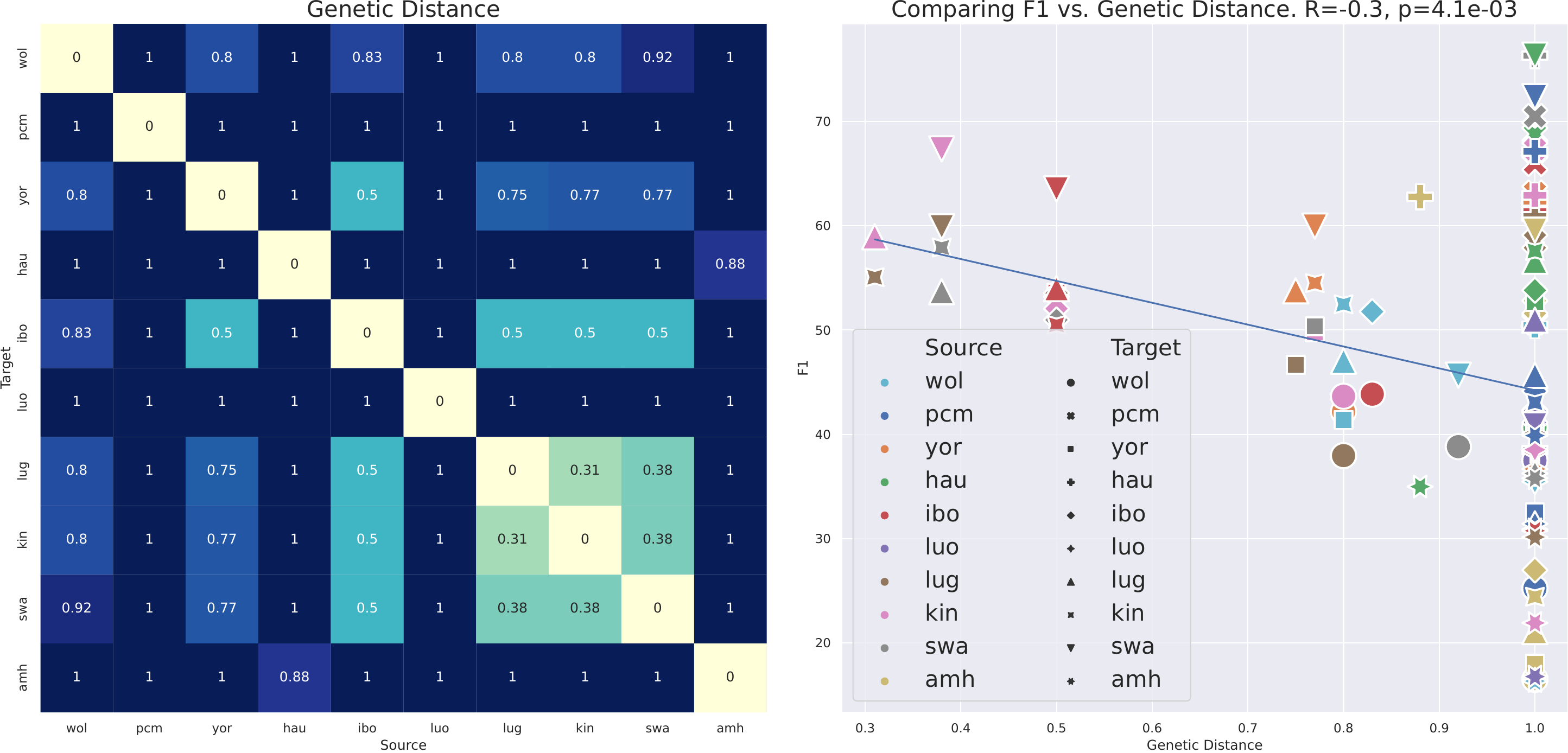}
  \caption{Genetic Distance}\label{fig:linetal:subplots_base_GENETIC}
  \centering\captionsetup{width=.95\linewidth}
  \includegraphics[width=0.9\linewidth]{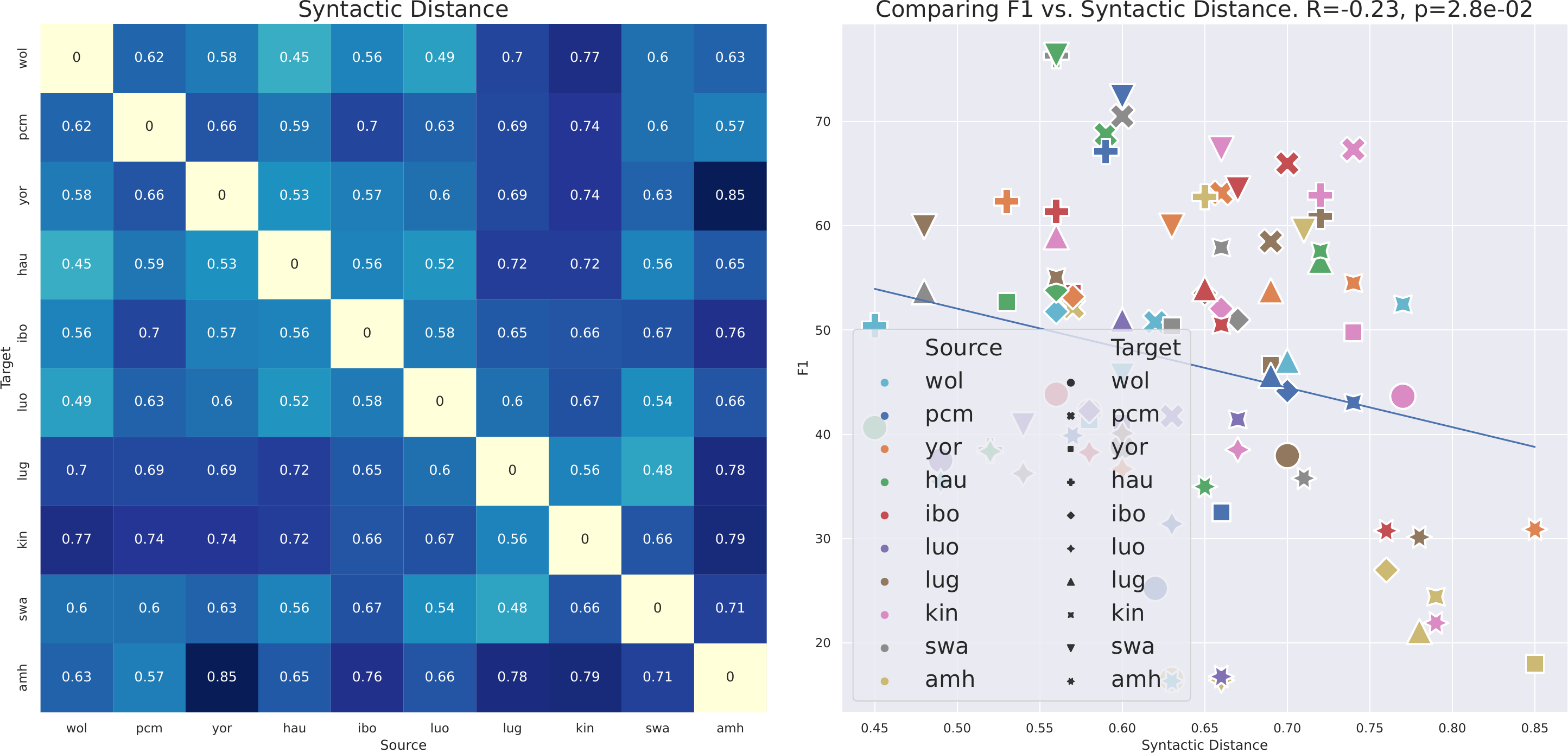}
  \caption{Syntactic Distance}\label{fig:linetal:subplots_base_SYNTACTIC}
  \centering\captionsetup{width=.95\linewidth}
  \includegraphics[width=0.9\linewidth]{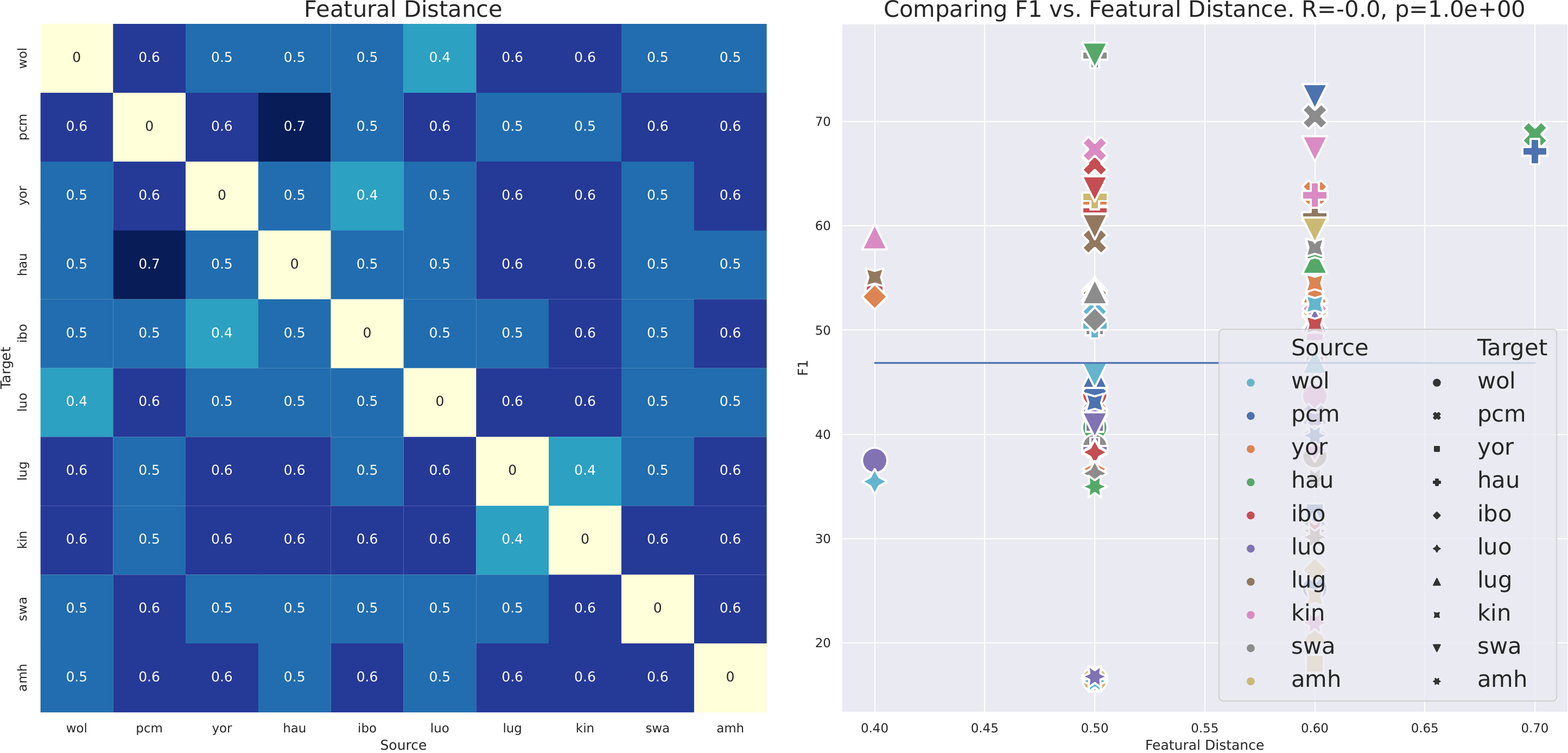}
  \caption{Featural Distance}\label{fig:linetal:subplots_base_FEATURAL}
  \end{minipage}
\end{figure*}

\begin{figure*}
  \centering
  \begin{minipage}[t]{1\textwidth}
  \centering\captionsetup{width=.95\linewidth}
  \includegraphics[width=0.9\linewidth]{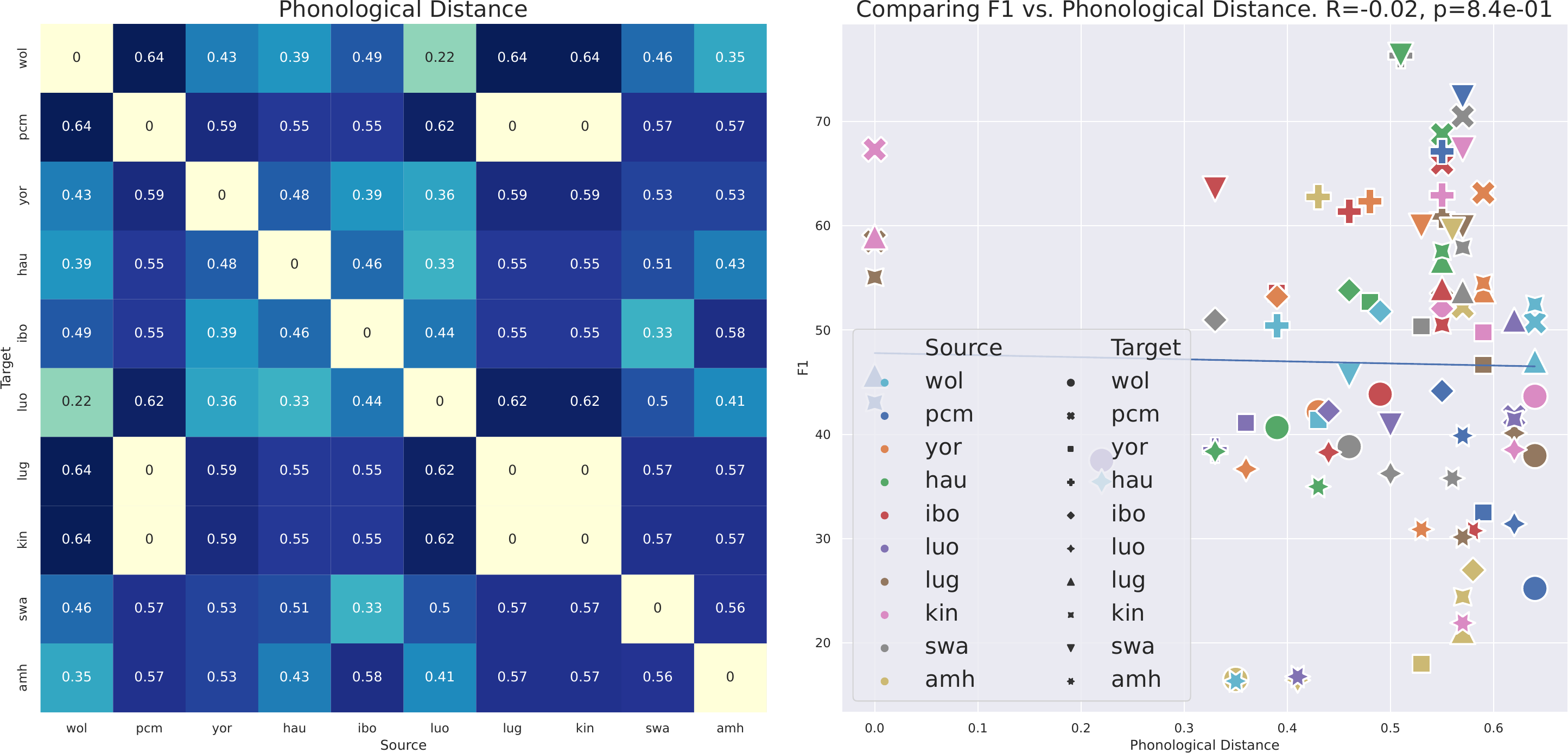}
  \caption{Phonological Distance}\label{fig:linetal:subplots_base_PHONOLOGICAL}
  \centering\captionsetup{width=.95\linewidth}
  \includegraphics[width=0.9\linewidth]{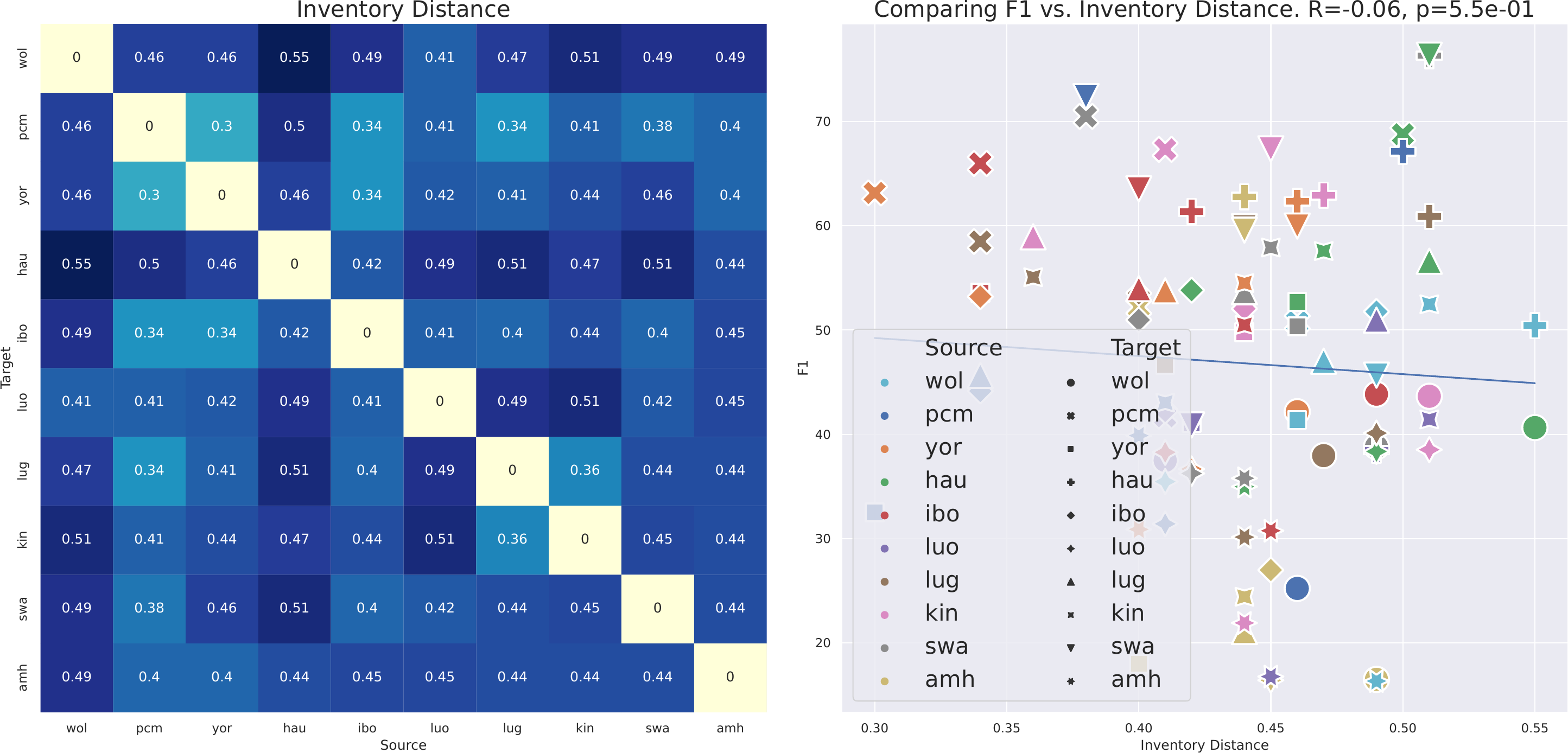}
  \caption{Inventory Distance}\label{fig:linetal:subplots_base_INVENTORY}
  \centering\captionsetup{width=.95\linewidth}
  \includegraphics[width=0.9\linewidth]{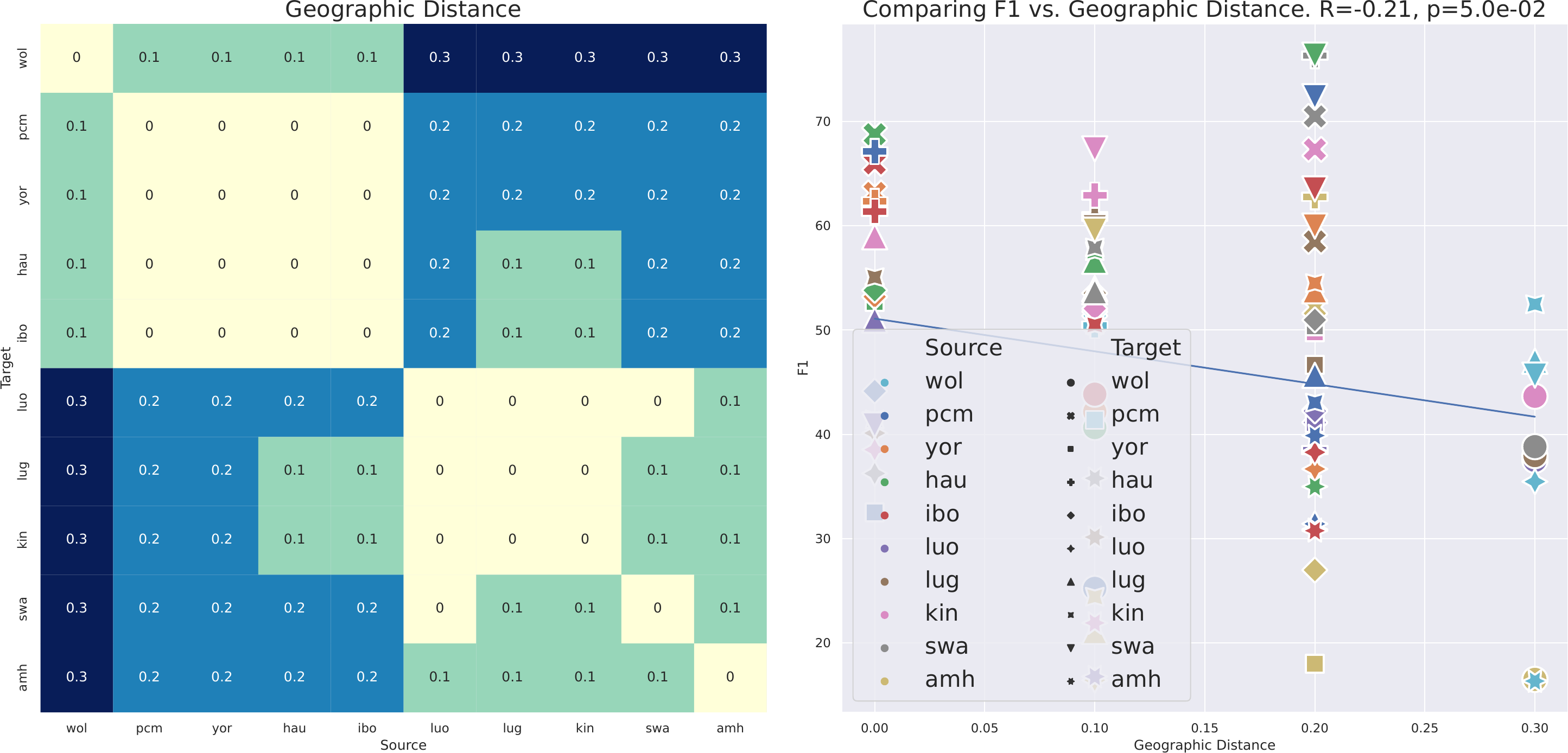}
  \caption{Geographic Distance}\label{fig:linetal:subplots_base_GEOGRAPHIC}
  \end{minipage}
\end{figure*}

  \end{document}